%% file: neurofusion.tex
\documentclass[acmtog]{acmart}

\acmSubmissionID{448}

\setcopyright{acmcopyright}\acmJournal{TOG}
\acmYear{2023}\acmVolume{40}\acmNumber{4}\acmArticle{56}\acmMonth{8} \acmDOI{10.1145/3450626.3459676}

\input{ourcommands}

\begin{document}
\title{MIPS-Fusion: Multi-Implicit-Submaps for Scalable and Robust Online Neural RGB-D Reconstruction}

\author{Yijie Tang$^{\ast}$}
\affiliation{%
  \institution{National University of Defense Technology}
  \country{China}
}
\author{Jiazhao Zhang$^{\ast}$}
\affiliation{%
  \institution{CFCS, Peking University}
  \country{China}
  \authornote{Joint first authors.}
}
\author{Zhinan Yu}
\affiliation{%
  \institution{National University of Defense Technology}
  \country{China}
}

\author{He Wang}
\affiliation{%
  \institution{CFCS, Peking University}
  \country{China}
}

\author{Kai Xu$^{\dagger}$}
\affiliation{%
  \institution{National University of Defense Technology}
  \country{China}
\authornote{Corresponding author: Kai Xu (kevin.kai.xu@gmail.com)}}

\begin{abstract}
\input{abstract}
\end{abstract}

\vspace{-18pt}
\ccsdesc[500]{Computing methodologies~Shape modeling}

\vspace{-26pt}
\keywords{Online RGB-D reconstruction, neural implicit representation, random optimization}
\vspace{-28pt}

\input{figure/teaser}

\maketitle

\input{intro}

\input{related}

\input{method}

\input{implement}

\input{results}

\input{conclusion}

\begin{acks}
We thank the anonymous reviewers for their valuable comments. We are grateful to Qi Wu for the fruitful discussions. This work was supported in part by the National Key Research and Development Program of China (2018AAA0102200), NSFC (62325211, 62132021).
\end{acks}

\bibliographystyle{ACM-Reference-Format}
\bibliography{neurofusion}

\end{document}

%% file: ourcommands.tex
\usepackage{mathtools}
\usepackage{amsmath}

\usepackage{amssymb}
\usepackage{amsfonts}
\usepackage{amsopn}
\usepackage{graphicx} 
\usepackage{textcomp}
\usepackage{xfrac}
\usepackage{bbm}
\usepackage{overpic}
\usepackage{subfig}
\usepackage{wrapfig}
\usepackage[ruled,vlined,linesnumbered]{algorithm2e}
\SetAlFnt{\small}
\SetAlCapFnt{\small}
\SetAlCapNameFnt{\small}
\SetAlCapHSkip{0pt}

\usepackage{multirow}

\usepackage{booktabs}
\usepackage{footnote}
\usepackage{paralist}
\usepackage{enumitem}
\usepackage{autobreak}
\usepackage{hyperref}
\usepackage{cleveref}

\usepackage{color}
\definecolor{green}{rgb}{0, 0.4, 0}
\definecolor{orange}{rgb}{0.8, 0.6, 0.2}
\definecolor{red}{rgb}{1.0, 0.0, 0.0}
\definecolor{teal}{rgb}{0.0, 0.4, 0.4}
\definecolor{purple}{rgb}{0.65,0,0.65}
\definecolor{saffron}{rgb}{0.95,0.75,0.2}
\definecolor{turquoise}{rgb}{0.0,0.5,0.5}
\definecolor{brown}{rgb}{0.5, 0.16, 0.16}

\usepackage{overpic}
\usepackage{currfile} 

\newlength\savedwidth

\newcommand{\rev}[1]{{\color{black}#1}}

\definecolor{lightgray}{rgb}{0.6, 0.6, 0.6}

\newcommand{\jiazhao}[1]{{\textcolor{black}{#1}}}

\newcommand{\yijie}[1]{{\textcolor{black}{#1}}}

\newcommand{\final}[1]{{\textcolor{black}{#1}}}
\newcommand{\ffinal}[1]{{\textcolor{black}{#1}}}

\newcommand{\Fig}[1]{Figure~\ref{fig:#1}}
\newcommand{\Eq}[1]{Eq.~(\ref{eq:#1})}

\newcommand{\Sec}[1]{Section~\ref{sec:#1}}

\newcommand{\Tab}[1]{Table~\ref{tab:#1}}

\usepackage[normalem]{ulem}

\newcommand{\hidecomment}[1]{}

\newcommand{\cF}{\mathcal{F}}
\newcommand{\cS}{\mathcal{S}}
\newcommand{\cP}{\mathcal{P}}

\newcommand{\cL}{\mathcal{L}}
\newcommand{\cB}{\mathcal{B}}
\newcommand{\cC}{\mathcal{C}}

\newcommand{\bR}{\mathbf{R}}
\newcommand{\bn}{\mathbf{n}}
\newcommand{\br}{\mathbf{r}}
\newcommand{\bo}{\mathbf{o}}
\newcommand{\bx}{\mathbf{x}}
\newcommand{\bc}{\mathbf{c}}

\newcommand{\bs}{\mathbf{s}}
\newcommand{\bv}{\mathbf{v}}
\newcommand{\bq}{\mathbf{q}}
\newcommand{\bp}{\mathbf{p}}
\newcommand{\bt}{\mathbf{t}}
\newcommand{\bz}{\mathbf{z}}

\newcommand{\bepsilon}{{\bf \epsilon}}
\newcommand{\bT}{\mathbf{T}}

\newcommand{\bM}{\mathbf{M}}

\newcommand{\tdC}{\tilde{C}}
\newcommand{\tdD}{\tilde{D}}
\newcommand{\tW}{\text{W}}

\DeclareMathOperator*{\argmin}{arg\,min}

\newcommand{\best}[1]{{$\mathbf{\color{blue} #1}$}}
\newcommand{\sbest}[1]{{$\mathbf{\color{green} #1}$}}

\usepackage{xspace}

\newcommand{\DSTUM}{\texttt{TUM RGB-D}\xspace}

\newcommand{\DSFCM}{\texttt{FastCaMo}\xspace}
\newcommand{\DSFCMS}{\texttt{FastCaMo-Synth}\xspace}
\newcommand{\DSFCMR}{\texttt{FastCaMo-Real}\xspace}
\newcommand{\DSFCML}{\texttt{FastCaMo-Large}\xspace}
\newcommand{\DSREPL}{\texttt{Replica}\xspace}
\newcommand{\DSSN}{\texttt{ScanNet}\xspace}


%% file: abstract.tex
We introduce MIPS-Fusion, a robust and scalable online RGB-D reconstruction method based on a novel neural implicit representation -- multi-implicit-submap. Different from existing neural RGB-D reconstruction methods lacking either flexibility with a single neural map or scalability due to extra storage of feature grids, we propose a pure neural representation tackling both difficulties with a divide-and-conquer design. In our method, neural submaps are incrementally allocated alongside the scanning trajectory and efficiently learned with local neural bundle adjustments. The submaps can be refined individually in a back-end optimization and optimized jointly to realize submap-level loop closure. Meanwhile, we propose a hybrid tracking approach combining randomized and gradient-based pose optimizations. For the first time, randomized optimization is made possible in neural tracking with several key designs to the learning process, enabling efficient and robust tracking even under fast camera motions. The extensive evaluation demonstrates that our method attains higher reconstruction quality than the state of the arts for large-scale scenes and under fast camera motions.

%% file: figure/teaser.tex
\begin{teaserfigure}
\begin{overpic}[width=1.0\textwidth,tics=5]{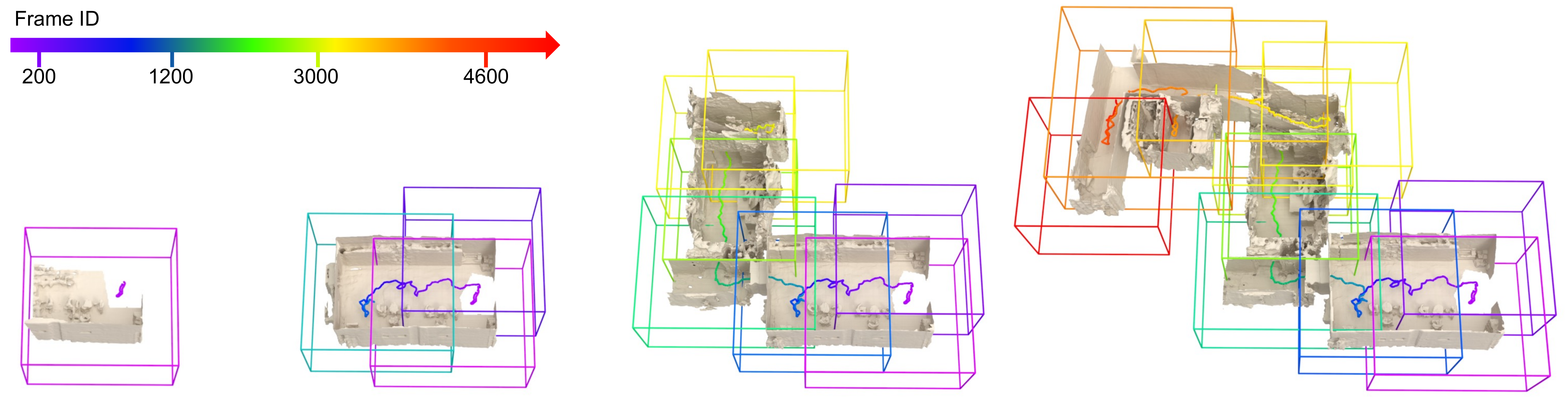}
\end{overpic}\vspace{-4pt}
\caption{
We introduce MIPS-Fusion, an online RGB-D reconstruction based on a novel neural implicit representation -- multi-implicit-submap. Neural submaps are allocated incrementally alongside the scanning trajectory, learned efficiently with local bundle adjustment, refined distributively with back-end optimization, and optimized globally with loop closure. The divide-and-conquer scheme attains both flexibility and scalability. We also propose a hybrid tracking approach where randomized optimization is made possible in the neural setting, enabling efficient and robust tracking even under fast camera motions.}
\label{fig:teaser}
\end{teaserfigure} 

%% file: intro.tex
\section{Introduction}
\label{sec:intro}
The recent decade has witnessed a proliferation of online dense reconstruction based on RGB-D cameras since the seminal work of KinectFusion~\cite{Newcombe2011,Izadi2011}.
Its core technique is simultaneous camera localization (tracking) and depth fusion (mapping).
Camera tracking has been a long-standing problem in 3D vision and robotics and has gained extensive research. Recently, the tracking robustness under fast camera motions has been drastically improved based on randomized optimization~\cite{zhang2021rosefusion,zhang2022asro}.
Contrary to the rapidly advanced frontiers of tracking, mapping received relatively less attention and the mainstream approach has been largely confined to volumetric~\cite{Curless96} and point-based fusion~\cite{keller2013real,Whelan2015} until recently when neural implicit mapping came along.

\input{figure/overview}

The learning of implicit representation of 3D objects and scenes is drawing increasing attention lately, with many powerful methods proposed and a recent climax reached by neural radiance fields (NeRF)~\cite{mildenhall2021nerf}.
Traditional dense reconstruction approaches adopting explicit volumetric representation suffer from the scalability issue. The storage cost makes it difficult to map a large scene such as a floor of a building of a moderate size.
Neural implicit representation seems a promising solution to scalable mapping since it encodes the scene with a compact, end-to-end learnable neural network. In several neural SLAM and RGB-D reconstruction works~\cite{sucar2021imap}, camera tracking can be jointly optimized with the neural map representation.
Despite the encouraging progress that has been made, however, neither the scalability of neural mapping nor the robustness of neural tracking is satisfactory to date.

Neural networks have the issue of limited learning capacity. Some recent works devise dense feature grids to mitigate the issue at the cost of cubic spatial complexity which is again hard to scale. NICE-SLAM~\cite{zhu2022nice} adopts hierarchical feature grids to improve scalability to some extent. We advocate the use of \emph{purely neural maps} and aim to fully exploit the potential of implicit representations. To mitigate the limited capacity issue without scarifying scalability, we propose incremental allocation and on-the-fly learning of multiple neural fields alongside the scanning trajectory (see \Fig{teaser}). Each neural field, coined \emph{neural submap}, governs a local subvolume and encodes the scene geometry and colors defined in its \emph{local coordinate frame}. The neural submaps are allocated incrementally and learned efficiently with a local bundle adjustment (BA). To achieve a smooth map transition, we ensure that adjacent neural submaps are spatially overlapping and updated with their shared keyframes jointly. This on-demand \emph{multi-implicit-submap} scheme allows a scalable reconstruction with rich local geometric details.

To achieve high tracking robustness, especially under fast camera motions, we propose a hybrid tracking scheme combining both gradient-based (GO) and randomized optimizations (RO). Realizing RO in the implicit setting is conceptually straightforward but computationally prohibitive since it needs to evaluate fitness for a sufficiently large number of hypothetic camera poses and each evaluation involves many times of network inference. To accelerate this process, we devise two key designs. \emph{First}, we propose a \emph{depth-to-TSDF loss} for which network inference is done only for points unprojected from the depth map and transformed by a hypothetic pose; no expensive volumetric depth rendering is needed as in the previous works~\cite{sucar2021imap,zhu2022nice}. Meanwhile, this loss is differentiable and admits GO. This allows for a scheduled optimization scheme: RO is used in early iteration steps to obtain a good initialization, which is followed by GO-based refinements. We can optionally optimize a photometric loss based on RGB rendering and GO when the RGB observations are reliable (e.g., texture-rich and blur-free). \emph{Second}, to further speed up the fitness evaluation, we opt for a \emph{light-weight network for classification-based TSDF prediction} trained to output a probabilistic distribution over a discrete set of distances. This makes our neural submaps easier to learn. More importantly, the epistemic uncertainty of TSDF classification can be used to build a weighted fitness for improved tracking accuracy.

The on-the-fly allocation of multiple neural submaps facilitates distributed map refinement. The submap corresponding to the current keyframe, referred to as \emph{active submap}, is usually insufficiently trained for the sake of maintaining realtime framerate. To this end, in parallel to the online updating of the active submap, we fine-tune the \emph{inactive submaps} in a separate thread using denser sampling of keyframes and depth pixels. Furthermore, our method also supports \emph{loop closure of neural submaps}. Once a non-trivial loop is detected, we perform a submap-level BA to jointly optimize the poses of all submaps in the loop. Since our neural submaps are defined in their local coordinate frames, map adjustment can be realized efficiently by transforming the submaps which is much faster than neural updating of learned submaps. See \Fig{overview} for an overview.

The design philosophy of our approach is a divide-and-conquer mapping scheme for flexibility and scalability, together with a hybrid tracking scheme for efficiency and robustness, both enabled by lightweight neural representations assisted with an easier task of classification. We have evaluated our method on several public benchmarks and a newly introduced dataset of large-scale scenes. On all benchmarks, our method outperforms the state-of-the-art neural SLAM/reconstruction methods. It also successfully reconstructs the challenging sequences with fast camera motions on which all previous neural methods failed.
In summary, the contributions of our work include:

\begin{itemize}
	\item We propose a purely neural mapping approach which achieves scalable dense RGB-D reconstruction through incrementally allocating and on-the-fly learning multiple neural submaps. 
 
	\vspace{7pt}\item We propose a robust neural tracking method which works well for fast camera motions via combining gradient-based and randomized optimizations in the neural representation.

    \vspace{7pt}\item \final{Our multi-implicit-submap approach supports parallel finetuning of submaps and, for the first time, realizes loop closure in neural mapping with submap-level BA.}
 
\end{itemize}

%% file: figure/overview.tex
\begin{figure*}[t]
\centering
\begin{overpic}
[width=\linewidth]
{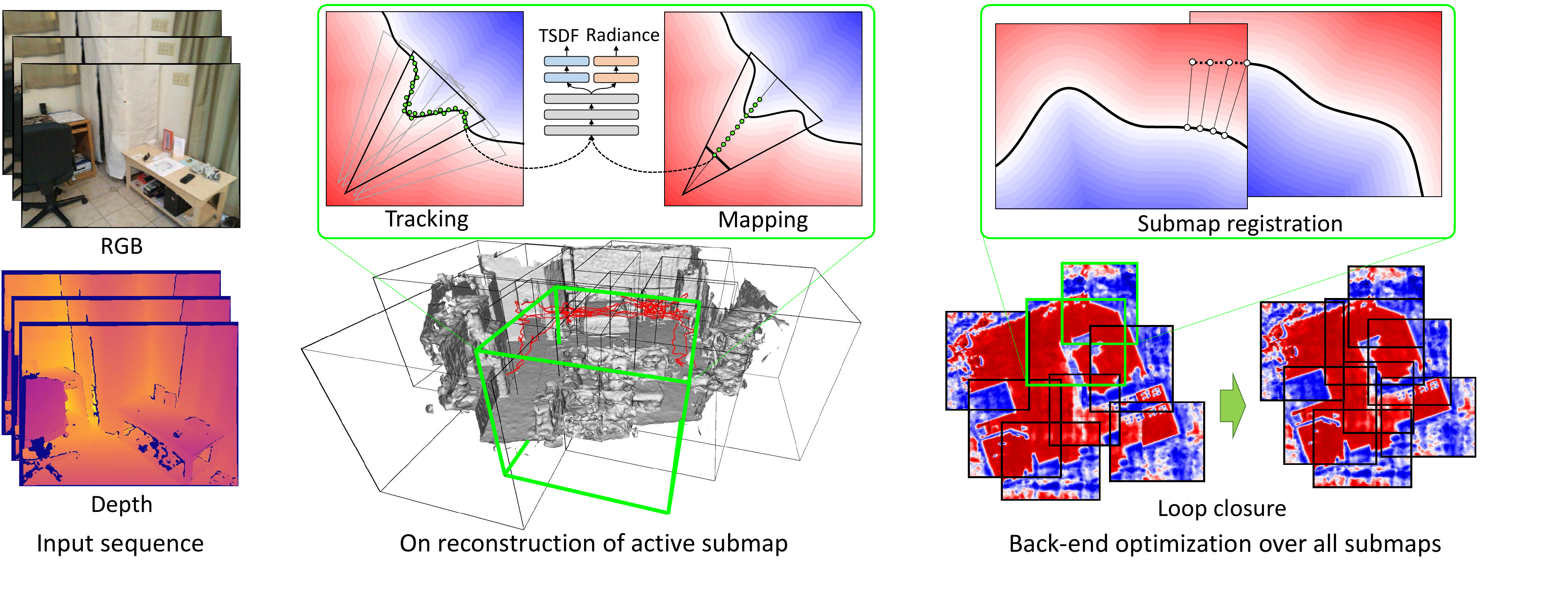}
\end{overpic}
\caption{
Method overview. Our method is comprised of online reconstruction of active submap based on neural tracking and mapping and back-end optimization over all inactive submaps based on intra-submap refinement and and inter-submap registration. The latter facilitates submap-level loop closure.
}
\label{fig:overview}
\end{figure*}

%% file: related.tex

\section{Related work}
\label{sec:related}
We will focus on works on \emph{dense SLAM and online RGB-D reconstruction}  and review them in terms of mapping and tracking separately covering both traditional and neural approaches, followed by a discussion on loop closure in the same context. 

\paragraph{Mapping.}
Since the seminal work of DTAM~\cite{newcombe2011dtam}, dense SLAM has been extensively studied over the years (see the survey by~\cite{taketomi2017visual}).
Taking advantage of RGB-D cameras, KinectFusion~\cite{Newcombe2011,Izadi2011} achieves the first online RGB-D reconstruction via realizing real-time volumetric depth fusion~\cite{Curless96}.
In order to handle larger environments, spatial hierarchies~\cite{chen2013scalable} and hashing schemes~\cite{Niessner2013,Kahler2015} have been proposed.
\final{Some recent works propose to learn depth map fusion to account for fusion errors~\cite{weder2020routedfusion, Cao2018RealtimeHT}, handle outliers~\cite{weder2021neuralfusion}, or preserve details~\cite{li2022bnv,lionar2021neuralblox}.
Another line of works adopt point-, surfel-~\cite{Whelan2012,keller2013real,henry2014rgb, Whelan2016ElasticFusionRD,wang2019real,pradeep2013monofusion, liu2016closed} which leads to better mapping scalability but produces lower map density.}
\ffinal{Recently, ~\citeN{Xu2022HRBFFusionA3} propose a unique mapping scheme based on on-the-fly implicits of Hermite Radial Basis Functions (HRBFs) demonstrating good accuracy and robustness of RGB-D reconstruction.}

Another approach to scalable mapping is to represent the global map as a combination of submaps, which dates back at least to the Atlas framework~\cite{bosse2003atlas}. The existing works that utilize explicit TSDF subvolumes to maintain map consistency can be largely classified into two categories, i.e., those which attempt to partition space and minimize overlap between subvolumes~\cite{henry2013patch,kahler2016real} and those which do not partition space~\cite{fioraio2015large,millane2018c}.

Neural implicit representation offers new opportunities for scalable mapping, taking advantage of the expressiveness and compactness of learned geometric priors.
CodeSLAM~\cite{bloesch2018codeslam} trains an encoder-decoder network to embed depth maps as low-dimensional codes which can be used to optimize key-frame poses.
DI-Fusion~\cite{huang2021di} proposes to learn geometric priors to embed 3D points in a low-dimensional latent space which can then be decoded into SDF values. Such learned geometric prior is, however, inaccurate in handling complex geometric details.
Recently, iMAP trains an implicit network~\cite {sucar2021imap} online to represent a scene. Several careful designs are made to attain a good trade-off between compactness and accuracy of mapping. Inspired by that, iSDF~\cite{ortiz2022isdf} learns to map with a neural SDF with novel self-supervision and sampling strategies. 
\citeN{azinovic2022neural} propose to represent scene surface using an implicit TSDF and incorporate this representation in the NeRF framework for rendering-based learning.
Block-NeRF~\cite{tancik2022block} scales NeRF to render city-scale scenes spanning multiple blocks but not support online reconstruction.

Observing that the prior works such as iMAP use a single MLP to represent the entire scene, which can only be updated globally and hence suffers from the forgetting issue when scanning a large scene, the following works NICE-SLAM~\cite{zhu2022nice} and Vox-Fusion~\cite{yang2022vox} propose a hybrid representation which combines multi-level grid-based features and a neural decoder, inspired by several recent works~\cite{liu2020neural,li2022vox,peng2020convolutional,sun2022direct}. The learnable grid-based features can be seen as a ``spatially distributed network'' with immense representation capacity. The decoder can be either pretrained \emph{a priori} or learned on-the-fly. \citeN{rosinol2022nerf} propose a geometric and photometric 3D mapping pipeline from monocular images based on hierarchical volumetric neural radiance fields. \final{Recently, \citeN{wang2023co} adopt parametric encoding to accelerate learning convergence based on the multiresolution hash encoding~\cite{muller2022instant} on NeRF. More recently, \citeN{johari2023eslam} propose a new scene representation consisting of multi-scale axis-aligned perpendicular feature planes (tri-plane features).}

Our method differs from the existing works in that it utilizes multiple MLPs to jointly represent the scene. The neural submaps can be learned and refined independently, achieving a balance of expressiveness, compactness, and flexibility.


\paragraph{Tracking.}
Regarding camera tracking, KinectFusion and DTAM estimate poses for the input depth maps using frame-to-model alignment based on point-to-plane ICP. To improve robustness, many works further adopt photometric and/or feature-based tracking~\cite{Whelan2015,Dai2017}. Bylow et al.~\shortcite{bylow2013real} realize a feature-free tracking through optimizing an objective defined with depth-to-TSDF conformation. While most state-of-the-art tracking approaches rely on gradient-based optimization, \citeN{zhang2021rosefusion} argue that gradient-based methods are brittle when handling fast camera motions due to the high nonlinearity of large pose optimization. They propose ROSEFusion which minimizes a depth-to-TSDF objective similar to~\cite{bylow2013real} using randomized optimization and achieves highly robust camera tracking under fast motions. Later, the method is extended to realize depth-inertial odometry with even higher tracking robustness~\cite{zhang2022asro}.

Some methods include back-end optimizations such as bundle adjustment~\cite{schops2019bad,Dai2017} and pose-graph optimization~\cite{kahler2016real,kerl2013dense} to improve tracking accuracy. These back-end optimizations are typically time-consuming and therefore conducted only for keyframes and invoked sparsely in time. Back-end optimization is also used for loop closure; see below.

In the context of neural SLAM and neural online reconstruction, camera tracking is solved either in a \emph{coupled} way based on inverse neural representation learning via differentiable volumetric rendering, or in a \emph{decoupled} manner where camera poses are optimized independently without relying on the neural representation.
Coupled approaches, such as iMAP, NICE-SLAM, and Vox-Fusion, adopt a render-and-compare paradigm where both RGB and depth maps are rendered and compared to the corresponding observations. Since volumetric rendering is expensive, it is done only for subsampled keyframes and pixels.
Generally speaking, while coupled solutions seem neat and more integrated, decoupled ones usually lead to more robust tracking results. Note, however, that decoupled approaches can employ not only traditional tracking methods (e.g.,~\cite{koestler2022tandem,chung2022orbeez}), but also neural tracking models. For example, iDF-SLAM~\cite{ming2022idf} learns a neural feature detector and DROID-SLAM~\cite{teed2021droid} learns a neural optical flow estimator for frame-to-frame registration.

Our tracking method belongs to \emph{coupled} approach since it runs completely on the neural representation. To attain high robustness, we, for the first time, integrate gradient-based and randomized optimizations in the neural setting, although not fully differentiable due to the randomized part. We use the same objective function for both optimization processes, facilitating a natural switching between the two for a scheduled optimization. This objective function also saves depth rendering. Combining an efficient GPU implementation, we realize the first neural tracker working under fast camera motions.

\input{figure/submap}

\paragraph{Loop closure.}
Loop closure is a classic technique in the back-end of SLAM systems. Deep learning has been mainly used in loop closure detection (e.g.,~\cite{merrill2018lightweight}). Once detected, traditional methods are used for global adjustment of the camera trajectory and the map. Regarding bundle adjustment, BA-Net~\cite{tang2018ba} proposes a learnable bundle adjustment layer which optimizes over a number of coefficients used to linearly combine a depth basis as well as the damping factor of the Levenberg-Marquardt algorithm. DROID-SLAM proposes a differentiable dense bundle adjustment layer which computes a Gauss-Newton update to camera poses and dense per-pixel depth to match the estimated optical flow. However, they were not shown to work for loop closure. In fact, adjusting maps, especially neural maps, is much harder than camera trajectories. \citeN{yuan2022algorithm} propose an algorithm for the $SE(3)$-transformation of neural implicit maps for remapping in loop closure. We did not follow this method since our submaps can be updated locally and transformed globally. In fact, multi-submap adjustment admits more DoFs than $SE(3)$ transformations.

%% file: figure/submap.tex
\begin{figure}[t]
\centering
\begin{overpic}
[width=\linewidth]
{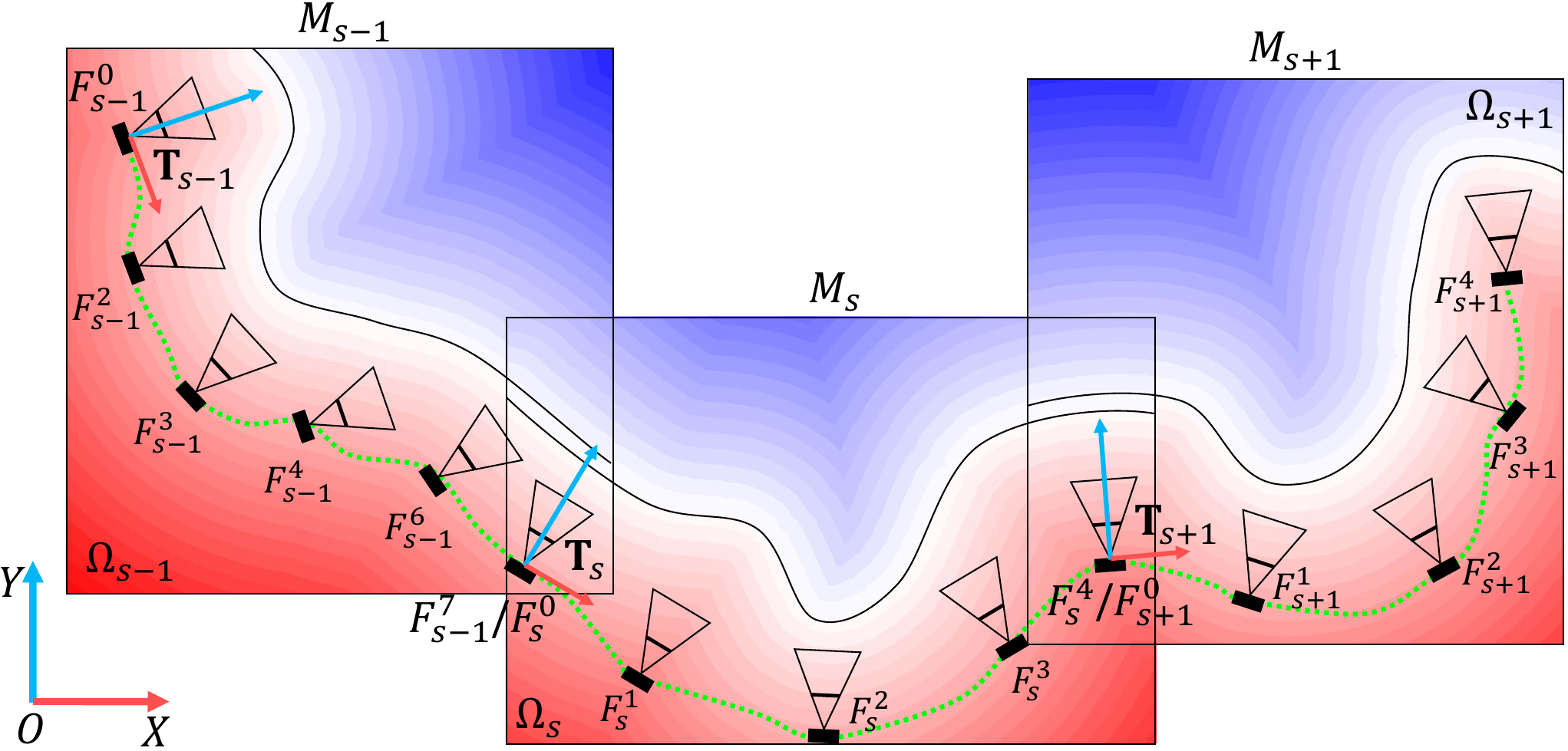}
\end{overpic}
\caption{
The multi-implicit-submap representation for online RGB-D reconstruction. Three submaps $M_{s-1}$, $M_{s}$ and $M_{s+1}$ with their subvolumes and keyframes are shown. For each submap, the first keyframe is the anchor at which the local coordinate frame of the submap is defined. Adjacent submaps share at least one keyframe, e.g., $F_{s-1}^7$ of $M_{s-1}$ is $F_{s}^0$ of $M_{s}$.
}
\label{fig:submap}
\end{figure}

%% file: method.tex
\section{Method}
\label{sec:method}

The input to online reconstruction is an RGB-D sequence $\{C_t, D_t\}_{t=0:T}$ ($C$ and $D$ are color and depth images, respectively) captured by an RGB-D camera and the output is a surface reconstruction of the scene being captured as well as a trajectory of 6DoF camera poses, $\{[\bR_t|\bt_t]\}_{t=0:T}$ ($[\bR|\bt] \in SE(3)$ represents a 6D camera pose in the world coordinate frame). Our method is built upon the neural mapping framework of iMAP~\cite{sucar2021imap}.
The key problem of online neural RGB-D reconstruction is the joint optimization of the neural map and the 6D camera pose of each frame.
\input{figure/pipeline}

\Fig{overview} gives an overview of our method. In this section, we first introduce our multi-implicit-submap representation (\Sec{rep}). Based on the representation, we describe the optimization losses used for mapping and tracking (\Sec{opt}). We then elaborate on the optimization processes for camera poses (\Sec{tracking}) and neural maps (\Sec{mapping}). Finally, we discuss the back-end optimization together with loop closure (\Sec{backend}).
\Fig{pipeline} shows our system pipeline which will be elaborated below.

\input{rep}

\input{loss}

\input{tracking}

\input{mapping}

\input{loop}

%% file: figure/pipeline.tex
\begin{figure}[t]
\centering
\begin{overpic}
[width=\linewidth]
{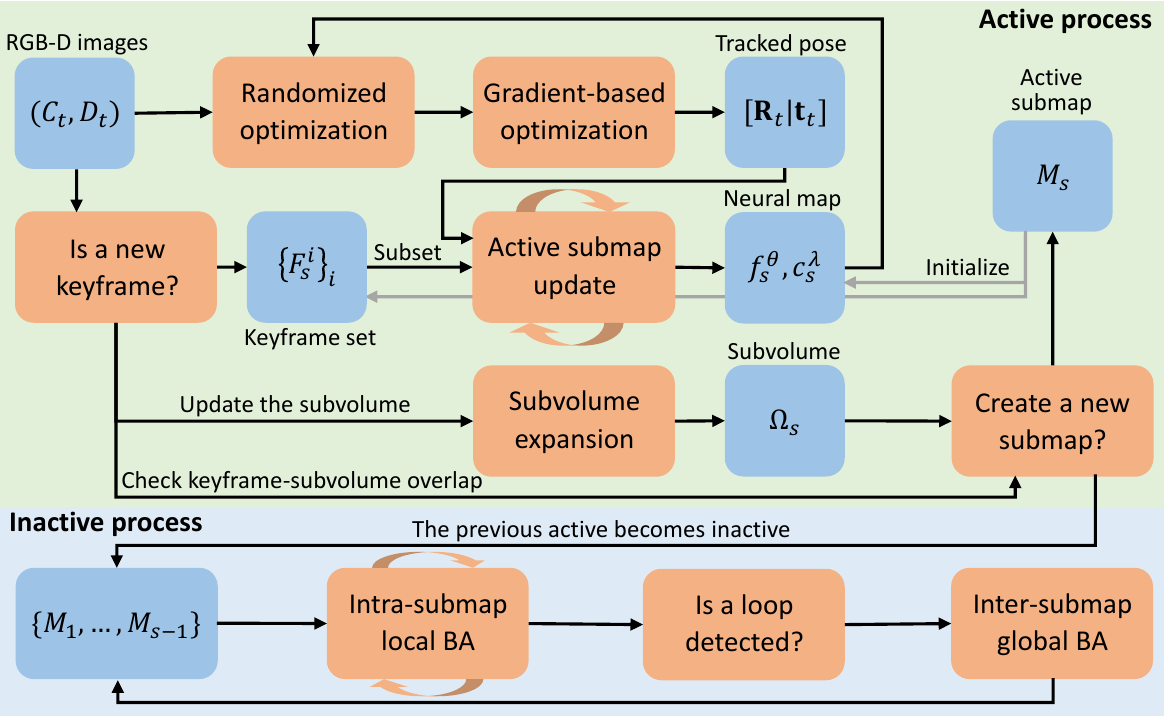}
\end{overpic}
\caption{
System pipeline. Our method runs two processes in parallel. The active process works on active submap and performs tracking of the current RGB-D frame, selection of keyframes, and mapping based on both the current frame and a subset of the keyframes. The subvolume of the active submap expands as new keyframes comes in. When a new active submap is created, the previous active submap becomes inactive. Back-end optimization is performed and loop closure conducted whenever available.
}
\label{fig:pipeline}
\end{figure}

%% file: rep.tex
\subsection{Multi-Implicit-Submap Representation}
\label{sec:rep}

Our scene representation is a sequence of $S$ neural submaps $\{M_s\}_{s=1:S}$ allocated alongside the scanning trajectory. Each submap is a tuple $M_s=\langle f^\theta_s, c^\lambda_s, \bT_s, \cF_s, \Omega_s \rangle$, where $f^\theta_s$ is the truncated signed distance function (TSDF) and $c^\lambda_s$ the radiance field, both implemented as a multi-layer perceptron (MLP) and parameterized by $\theta$ and $\lambda$, respectively. \final{More advanced scene representation techniques~\cite{wang2023co, muller2022instant} could be used for enhanced reconstruction quality.}
While $c^\lambda_s$ is trained as usual, $f^\theta_s$ is learned as a classifier to facilitate fast mapping and tracking. Both the two functions are defined in the local coordinate frame of the submap $M_s$ with $\bT_s$ being the global pose of the submap in the world coordinate frame.
$\cF_s$ is the set of keyframes associated to $M_s$. The global pose of the first keyframe $F_s^0 \in \cF_s$ is set to the submap pose $\bT_s$. Therefore, $F_s^0$ is also referred to as the anchor keyframe of $M_s$.
$\Omega_s$ is the axis-aligned cuboid subvolume that $M_s$ governs. See \Fig{submap} for illustration.

\input{figure/network}

Learning local neural map functions gains flexibility such that each submap can be transformed as a whole for efficient global alignment of submaps in loop closure. Meanwhile, local functions are generally easier to learn due to the low data bias caused by localized data distributions in local coordinate frames.

\paragraph{Classification-based neural TSDF}
Given a 3D point defined in the world coordinate frame and located in the subvolume of $M^s$, i.e., $\bx^\tW\in \Omega^s$, its TSDF value by $M^s$ can be computed as $\psi_s(\bx^s)=f^\theta_s(\bx^s)$, where $\bx^s=\bT^{-1}_s \bx^\tW$ is the point transformed into the local coordinate frame of $M^s$; see \Fig{network}. 
We use a sinusoidal positional encoding to encode the 3D position before feeding it into the neural network~\cite{mildenhall2021nerf}.
For a point located in the overlapping area of two subvolumes, e.g., $\bx^W\in \Omega^s \cap \Omega^t$, its global TSDF value can be evaluated as a weighted combination of the local TSDF values given by the corresponding two submaps:
\begin{equation}\label{eq:tsdf-blend}
  \psi(\bx^\tW)=\frac{w_s(\bx^s) \psi_s(\bx^s) + w_t(\bx^t) \psi_t(\bx^t)}{w_s(\bx^s)+w_t(\bx^t)},
\end{equation}
where $\bx^\ast=\bT^{-1}_\ast \bx^\tW$ and the weight is $w_\ast=\frac{1}{h_\ast(\bx^\ast)^2}$ with $h_\ast$ being the uncertainty of TSDF prediction by the submap (see below).

%
%
\rev{In particular, we choose $5$ signed distance values uniformly from the interval $[-\tau, \tau]$,
i.e., $\{\ell_1=-\tau, \ell_2=-\frac{\tau}{2}, \ell_3=0, \ell_4=\frac{\tau}{2}, \ell_1=\tau\}$,
where $\tau=0.1$m is the truncation distance.
Given a point $\bx^s$, $f^\theta_s$ outputs a normalized 5D vector $\bz=(z_i)_{i=1,\ldots,5}$ corresponding to the five distance values with $z_i$ indicating how probable $\bx^s$'s SDF value is close to $\ell_i$. We can then approximate $\bx^s$'s SDF value with a soft argmax:
\begin{equation}\label{eq:sdf-softmax}
   \psi_s(\bx^s)=\sum_{i=0}^{5}\frac{e^{\beta z_i}}{\sum_{i=1}^{5}e^{\beta z_i}}{\ell_i}, 
\end{equation}
where we use a coldness parameter $\beta=10$.}
\final{With soft argmax, our method obtains gradients from the training losses, facilitating effective learning of submaps. Moreover, the coldness parameter controls the smoothness of the probabilities over multiple classes, leading to a better approximation of SDF distributions.}
\Fig{plot-converge} shows that our classification-based TSDF prediction converges faster and hence learns faster than regression-based one.

Defining the probability distribution over the five classes as $p_i=\frac{e^{\beta z_i}}{\sum_{i=1}^{5}e^{\beta z_i}},i=1,\ldots,5$, we can measure the uncertainty of the TSDF classification as the Shannon entropy: $h_s(\bx^s)=-\sum_{i=1}^{5}{p_i \log p_i}$.
\final{The plots in \Fig{plot-uncertainty} demonstrate that the uncertainty measurement is useful in filtering points with inaccurate TSDF predictions, and is insensitive to class count.}

\paragraph{Neural radiance field}
In addition to the neural geometry representation, we also learn for each submap a neural appearance representation~\cite{mildenhall2021nerf}, $c^\lambda_s$, for optimizing mapping and tracking with photometric losses. Similar to~\cite{sucar2021imap}, we omit the encoding of view directions since we are not interested in modeling view-dependent effects such as specularities.
Implemented also with MLPs, it takes as input a 3D position (after sinusoidal encoding) $\bx^s$ and regresses a radiance value $c^\lambda_s(\bx^s)$ as output.
This simplification also makes $c^\lambda_s$ light-weight and faster to learn.

\input{figure/plot_converge}

\paragraph{Color and depth map rendering}
We render a color image as a weighted sum of radiance values of points $\bq=\bo+d_p(\bq)\bv_p$ sampled along the ray $\bv_p$ shooting from the camera center $\bo$ to an image pixel $p$, with $d_p(\bq)$ being $\bq$'s depth.
The weights are computed directly from signed distance values as the product of two sigmoid functions~\cite{azinovic2022neural}:
\begin{equation}\label{eq:render-weight}
  \omega_p(\bq)=\sigma\left(\frac{\psi(\bq)}{\tau}\right)\sigma\left(-\frac{\psi(\bq)}{\tau}\right).
\end{equation}
The color along of pixel $p$ is approximated as a weighted sum of radiance sampled on ray $\bv_p$ \emph{within only the truncation region}:
\begin{equation}\label{eq:render-rgb}
  \tdC(p)=\frac{1}{\sum_{\bq\in \cS_p^\text{tr}}{\omega_p(\bq)}}\sum_{\bq\in \cS_p^\text{tr}}{\omega_p(\bq) c^\lambda_s(\bq, \bv_p)},
\end{equation}
where $\cS_p^\text{tr}$ is the set of sampled points in truncation region. Sampling only within truncation region leads to a much faster rendering with limited quality degrading since the weights drop quickly to zero outside the truncation according to \Eq{render-weight}.
The simplifications on rendering is fine to our task.
Depth can be rendered similarity:
\begin{equation}\label{eq:render-d}
  \tdD(p)=\frac{1}{\sum_{\bq\in \cS_p^\text{tr}}{\omega_p(\bq)}}\sum_{\bq\in \cS_p^\text{tr}}{\omega_p(\bq) d(\bq)}.
\end{equation}
Note that, however, the depth rendering is used only for visualizing the learned geometry; neither our mapping or tracking involves depth rendering loss due to its high computational cost.

%% file: figure/network.tex
\begin{figure}[t]
\centering
\begin{overpic}
[width=0.9\linewidth]
{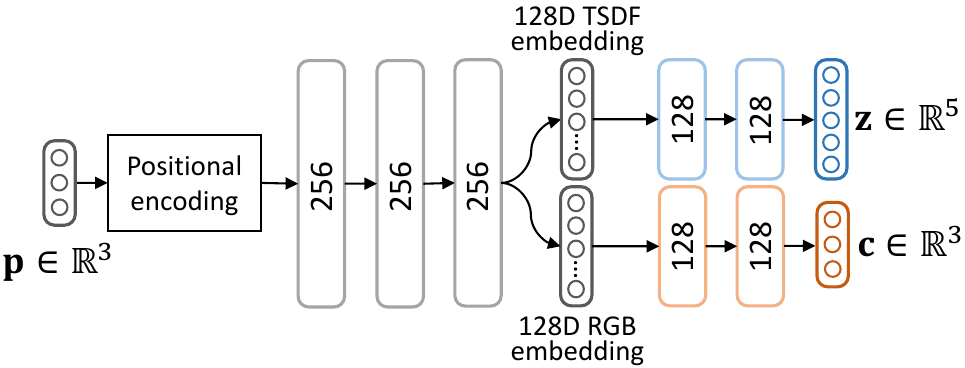}
\end{overpic}
\caption{
The network architecture of SDF and color prediction.
}
\label{fig:network}
\end{figure}

%% file: figure/plot_converge.tex
\begin{figure}[t]
\centering
\begin{overpic}
[width=\linewidth]
{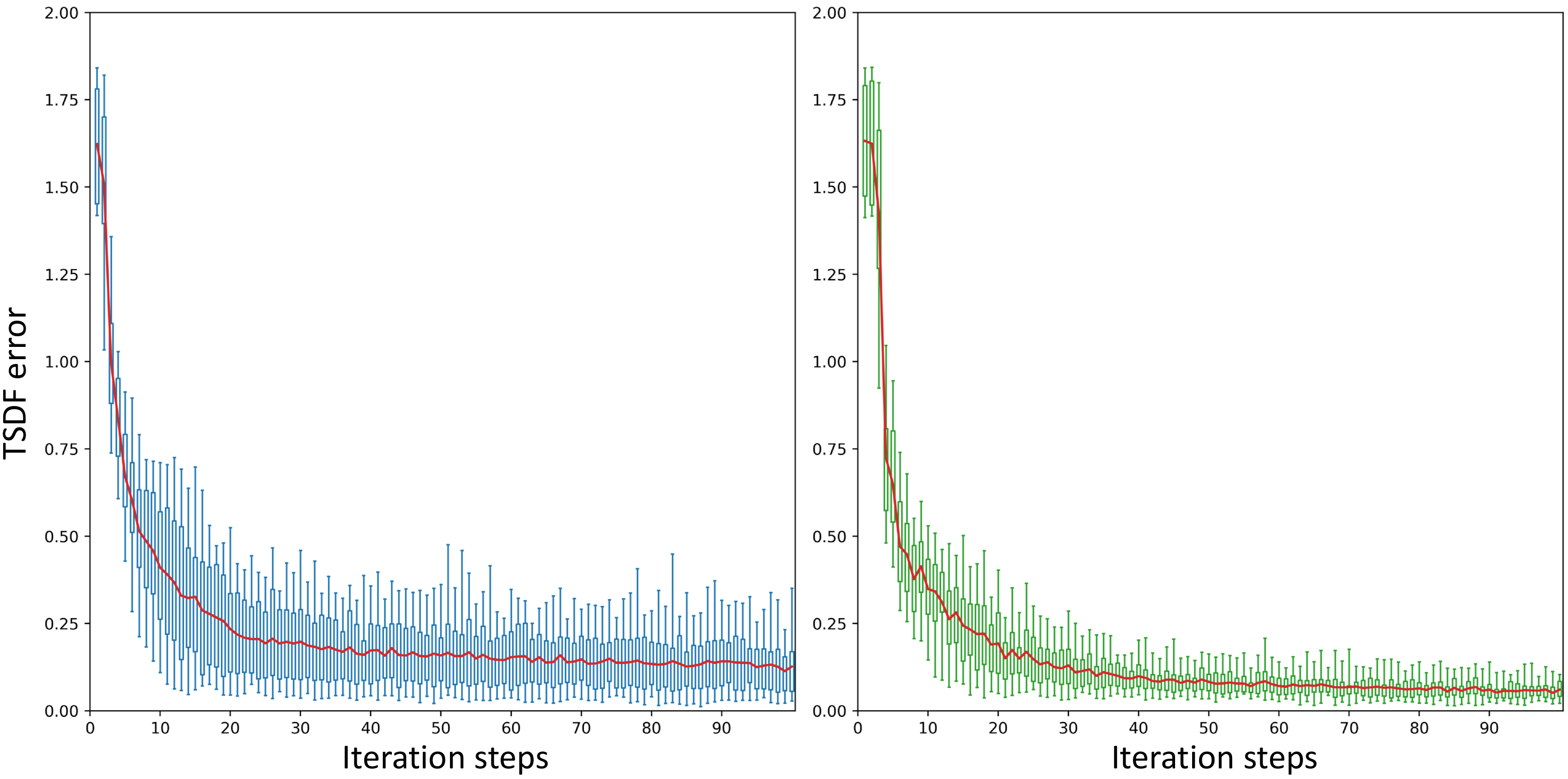}
    \put(21,50){\footnotesize (a) Regression.}
    \put(68,50){\footnotesize (b) Classification.}
\end{overpic}
\caption{
Convergence rate (learning speed) comparison between regression- and classification-based TSDF prediction.
}
\label{fig:plot-converge}
\end{figure}

%% file: loss.tex
\input{figure/plot_uncertainty}

\subsection{Optimization Losses for Mapping and Tracking}
\label{sec:opt}
To realize mapping and tracking, we optimize the neural scene representations together with the keyframe poses through minimizing different combinations of four different losses. The four losses include (1) a depth-to-TSDF loss $\cL_\text{d2t}$ for imposing the confirmation between the posed depth map and the learned TSDF, (2) a TSDF truncation-region loss $\cL_\text{tr}$ for learning the SDF values within the truncation region, (3) a TSDF free-space loss $\cL_\text{fs}$ for learning the truncation of TSDF on the visible side of the surface within the viewing frustum, and (4) an RGB rendering loss $\cL_\text{rgb}$ for enforcing photometric consistency.

\ffinal{While $\cL_\text{d2t}$ is used for tracking (RO and GO), $\cL_\text{tr}$, $\cL_\text{fs}$ and $\cL_\text{rgb}$ are used for both mapping and GO-based pose optimization.}
While the latter three losses are commonly seen in recent works, the depth-to-TSDF loss is new to neural SLAM and we show through evaluations that it is highly effective and efficient for tracking optimization.
We do not use the depth rendering loss of~\cite{yang2022vox} since it has been encompassed by $\cL_\text{tr}$ and $\cL_\text{fs}$.

\paragraph{Depth-to-TSDF loss}
Given the depth image of the current frame $D_t$ and the current neural submap $f^\theta_s$, our task is to compute the 6-DoF camera pose of the current frame in the world coordinate frame $[\bR_t|\bt_t]\in SE(3)$
while optimizing the $f^\theta_s$ with the 3D information of the posed depth map.
To this end, we define a frame-to-model error metric to measure the fitness of how well $D_t$ ``fits into'' the TSDF under pose $[\bR_t|\bt_t]$~\cite{bylow2013real,zhang2021rosefusion}.
For each pixel $p$ of $D_t$, we can compute based on its depth $D_t(p)$ the corresponding 3D point $\bx_p$ in the camera coordinate frame of the current frame. We can then transform this point into the world coordinate frame:
\begin{equation}\label{eq:unproj}
\bx_p^\tW = \bR_t \bx_p + \bt_t.
\end{equation}
We use the unprojected 3D points to query the TSDF map defined in the world coordinate system and obtain point-to-surface distances directly. If the camera pose is correct, it is expected that the point-to-surface distances of all unprojected 3D points should be zero.
%
Assuming that the depth measurements contain Gaussian noise and that all pixels are independent and identically distributed, the likelihood of observing depth image $D_t$ from camera pose $[\bR_t|\bt_t]$ is
\begin{equation}\label{eq:likelihood}
  p(D_t|\bR_t,\bt_t) \varpropto \prod_{p\in \cP}\exp\left(-\psi_s(\bT_s^{-1}(\bR_t\bx_p+\bt_t))^2\right),
\end{equation}
where $\cP$ is the set of sampled pixels. Our depth-to-TSDF loss is defined as the negative log-likelihood:
\begin{equation}\label{eq:d2tsdf-loss}
  \cL_\text{d2t}(\bR_t,\bt_t) = -\log{p(D_t|\bR_t,\bt_t)} = \sum_{p\in \cP}{\psi_s(\bT_s^{-1}(\bR_t\bx_p+\bt_t))^2}.
\end{equation}
The loss is used only for pose optimization of the current frame.


\paragraph{TSDF truncation-region loss}
The truncation-region loss is devised to supervise the MLP to output correct SDF values for points within the truncation region:
\begin{equation}\label{eq:tr-loss}
  \cL_\text{tr}(\Theta) = \frac{1}{|\cP|}\sum_{p\in \cP}{\frac{1}{|\cS_p^\text{tr}|}\sum_{\bq\in \cS_p^\text{tr}}{\left(\psi_s(\bT_s^{-1}\bq)-(D_t(p)-d_p(\bq))\right)^2}},
\end{equation}
where $\cS_p^\text{tr}$ is the set of points sampled on ray $\bv_p$ and within the truncation region. $D_t(p)-d_p(\bq)$ is the signed distance value of sample point $\bq$ 
\ffinal{with $d_p(\bq)$ being the sampled depth along ray $\bv_p$ of pixel $p$.
$\Theta$ may encompass the parameters of the TSDF $\theta$ and the camera pose, depending on whether the task is mapping or tracking.}
The predicted signed distance $\psi_s(\bT_s^{-1}\bq)$ is computed based on the output of $f^\theta_s$ according to \Eq{sdf-softmax}. This loss is used for optimizing the neural map parameters $\theta$. To ensure a meaningful uncertainty measurement of $f^\theta_s$'s output $(z_i)_{i=1,\ldots,5}$, we additionally minimize the following EMD-based distribution loss:
\begin{equation}\label{eq:tr-emd-loss}
  \cL_\text{tr-emd}(\Theta) = \frac{1}{|\cP|}\sum_{p\in \cP}{\frac{1}{|\cS_p^\text{tr}|}\sum_{\bq\in \cS_p^\text{tr}}{\sum_{i=1}^{5}z_i|i-y(\bq)|}},
\end{equation}
where $y(\bq)$ is the ground-truth label of TSDF classification at $\bq$.

\paragraph{TSDF free-space loss}
The free-space loss directs the neural map to output a value equal to the truncation value $\tau$ for the empty region in the visible side of the viewing frustum:
\begin{equation}\label{eq:fs-loss}
  \cL_\text{fs}(\Theta) = \frac{1}{|\cP|}\sum_{p\in \cP}{\frac{1}{|\cS_p^\text{fs}|}\sum_{\bq\in \cS_p^\text{fs}}{\left(\psi_s(\bT_s^{-1}\bq)-\tau\right)^2}},
\end{equation}
where $\cS_p^\text{fs}$ is the set of sample points in the free space of the visible side of ray $\bv_p$.
For free-space TSDF, a similar distribution loss $\cL_\text{fs-emd}$ is defined as in \Eq{tr-emd-loss}.

\paragraph{Remarks on the TSDF losses}
The depth-to-TSDF loss is estimated by direct point query and accounts only for 3D points unprojected from the depth map. This makes it much more efficient than volumetric rendering. Therefore, it is suited for depth-based tracking. The TSDF truncation-region and free-space losses concern about the full occupancy (geometry) information in the viewing frustum of a frame, which is thus well-targeted for the mapping task.

\paragraph{RGB rendering loss}
The RGB loss measures the squared differences between the rendered and the input (ground-truth) color images:
\begin{equation}\label{eq:rgb-loss}
  \cL_\text{rgb}(\Lambda) = \frac{1}{|\cP|}\sum_{p\in \cP}{\|\tdC(p)-C_t(p)\|},
\end{equation}
where $C_t(p)$ is the color at pixel $p$ of the input RGB image of the current frame and $\tdC(p)$ is defined in \Eq{render-rgb}.
$\Lambda$ may encompass the parameters of the radiance field $\lambda$ and the camera pose, depending on whether the task is mapping or tracking.

%% file: figure/plot_uncertainty.tex
\begin{figure}[t]
\centering
\begin{overpic}
[width=\linewidth]
{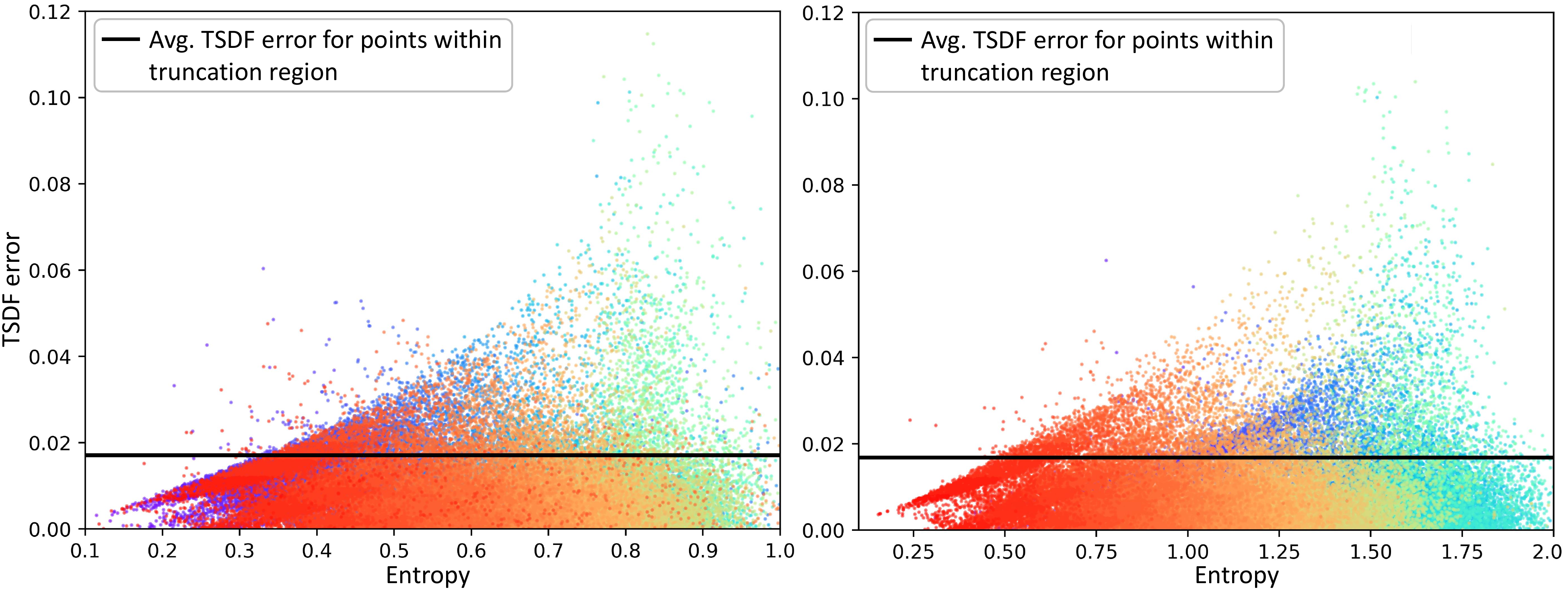}
    \put(20,38.5){\footnotesize (a) 5 classes.}
    \put(69,38.5){\footnotesize (b) 11 classes.}
\end{overpic}
\caption{
Scattered plots of uncertainty (Shannon entropy) of samples with different TSDF errors. The two plots are for TSDF classification against $5$ and $11$ classes, respectively. The colors of the dots indicate the TSDF values (red is large (free space), blue is small ($-\tau$) and green corresponds largely to the zero-level set). Points with larger TSDF errors generally have higher uncertainties and vice versa, making the uncertainty measurement useful in filtering points with inaccurate TSDF predictions. Note also that the majority of dots reside in the lower right instead of the upper left meaning that the entropy-based measure prefers false positive (FP) over false negative (FN). FN tends to trust predictions with higher TSDF error, which is more harmful to reconstruction accuracy.
}
\label{fig:plot-uncertainty}
\end{figure}

%% file: tracking.tex
\subsection{Tracking with Hybrid Optimization}
\label{sec:tracking}
Given the current RGB-D frame, we compute its camera pose by minimizing the depth-to-TSDF loss and the RGB rendering loss while keeping the current active submap fixed. To do so, we employ a hybrid optimization combining both
randomized optimization (RO) and gradient-based optimization (GO). The optimization is scheduled as follows. The RO is first performed for the depth-to-TSDF loss only. After a fixed number of RO iterations, the GO is invoked for both two losses for another fixed number of iterations.

\paragraph{Randomized pose optimization}
\input{figure/ro_vis}
The RO adopts the particle filter optimization (PFO) framework which samples and propagates a population of particles (candidate solutions) iteratively to make them cover the optimal solution as quickly as possible.
In our case, a solution is a 6DoF camera pose $\pi=(\bR, \bt)\doteq(q_x, q_y, q_z, x, y, z)$ with $q_x$, $q_y$ and $q_z$ being the imaginary part of the rotation quaternion and $\bt=(x,y,z)^T$.
The key to making PFO realtime capable is to \emph{pre-sample} a set of particles uniformly within the unit sphere in the 6D solution space, referred to as Particle Swarm Template (PST), instead of sampling the particles on the fly throughout the optimization. The PST is then moved and rescaled into a 6D ellipsoid over the optimization iterations to gradually cover the optimal solution.
Let us denote the PST at iteration $k$ as $\Pi_k$, which is parameterized by a center position $\bc$ and a vector of axis lengths (each for one of the six dimensions) $\br=(r_d)_{d=1:6}$.

In each iteration step $k$, we first evaluate the depth-to-TSDF loss $\eta$ for each particle $\pi_k^i=(\bR_k^i, \bt_k^i) \in \Pi_k$ based on the \emph{uncertainty-weighted} depth-to-TSDF loss:
\begin{equation}\label{eq:fitness}
  \eta(\pi_k^i)=\sum_{p\in \cP}{\frac{\psi_s(\bT_s^{-1}(\bR_k^i\bx_p+\bt_k^i))^2}{h_s(\bx_p)^2}},
\end{equation}
where $\bx_p$ is the unprojected 3D point corresponding to pixel $p$.
The uncertainty $h_s(\cdot)$ of TSDF prediction $\psi_s(\cdot)$ has been given in \Sec{rep}.
Among all the particles in $\Pi_k$, we collect those whose depth-to-TSDF loss is smaller than $\pi^\ast_{k-1}$ (the best solution in step $k-1$) into an Advantage Particle Set (APS).
%
The best state at the current step $k$ takes the centroid of the APS.
%
The PST is moved to be centered at the best state of the current step $\pi^\ast_{k}$.

To rescale the PST, we compute the axis lengths $\br_k$ of the current step as follows:
\begin{eqnarray}
  & \bv = \pi^\ast_k - \pi^\ast_{k-1},\label{eq:rsdir}\\
  & \hat{\br}_k = \eta(\pi^\ast_k) \frac{\bv}{\|\bv\|} + \bepsilon,\label{eq:rsinter}
\end{eqnarray}
where $\bv$ is an anisotropic attractor which drives the particles towards the best solution $\bs^\ast_k$ (the global best of the particle set). $\hat{\br}_k$ is the (interim) vector of axis lengths of $\Pi_{k}$, which is scaled by depth-to-TSDF loss $\eta(\bs^\ast_k)$ to gradually decrease the search range for stable convergence. $\bepsilon$ is a 6D vector of small numbers ($10^{-3}$) used to avoid degenerating PST. \final{Figure~\ref{fig:vis_ro} gives an illustration of randomized optimization.}
\final{The final shape of the PST is a blend between the current step axis lengths $\hat{\br}_k$ and previous step axis lengths $\br_{k-1}$:}
\begin{equation}\label{eq:momentum}
    \br_k = \alpha \br_{k-1} + (1-\alpha) \hat{\br}_k,
\end{equation}
where $\alpha=0.1$. The scaling factor is computed with $\br_k$ and $\br_{k-1}$.

\paragraph{Gradient-based pose optimization}
The GO phase minimizes the following loss over the input pixel batch set $\cB$:
\begin{equation}\label{eq:fs-loss}
  \cL(\bR_t,\bt_t) = \sum_{b\in \cB}{\cL_\text{d2t}^b(\bR_t,\bt_t) + \omega\cL_\text{rgb}^b(\bR_t,\bt_t)},
\end{equation}
where $\cL_\ast^b$ is the average loss over batch $b$ and $\omega=1$. We adopt the ADAM solver~\cite{kingma2014adam} with a learning rate of $10^{-2}$. 

%% file: figure/ro_vis.tex
\begin{figure}[t]
\centering
\begin{overpic}
[width=1\linewidth]
{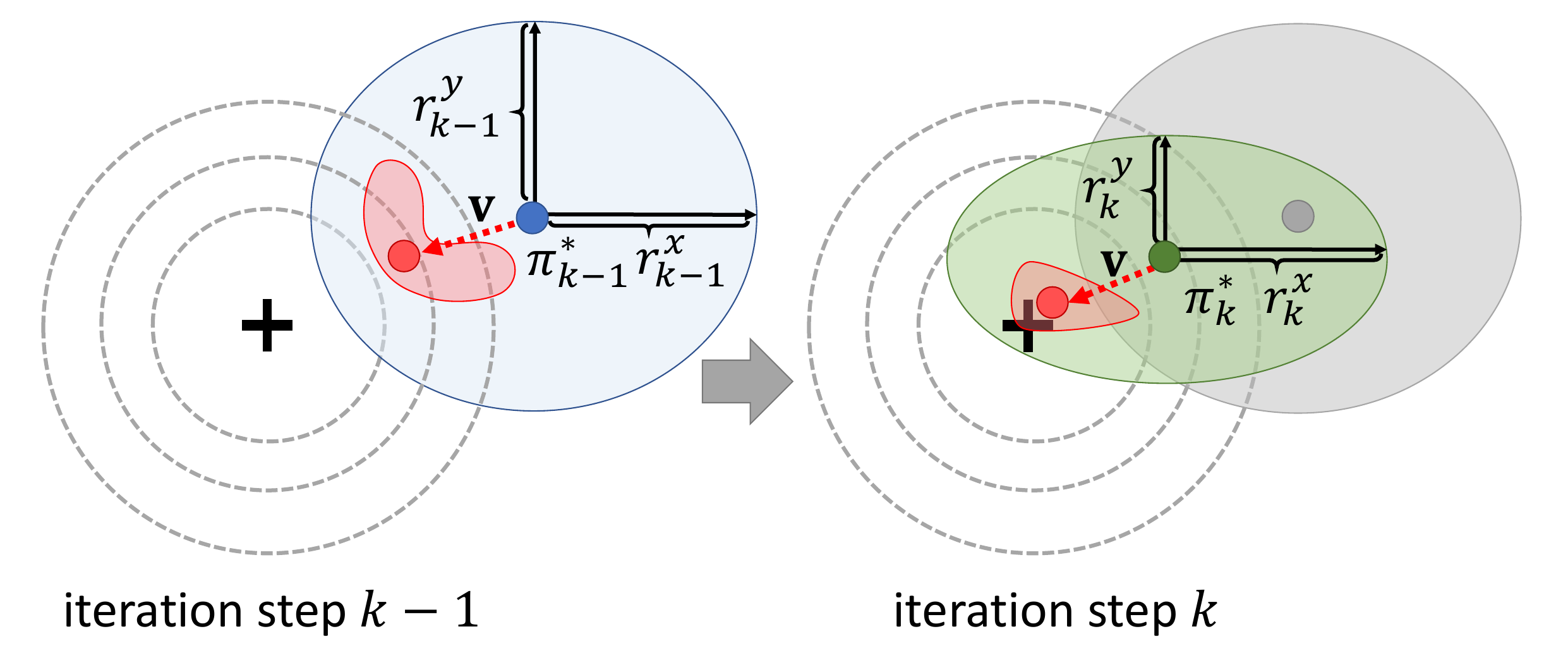}  
\end{overpic}
\caption{
\final{Moving and rescaling PST from iteration step $k-1$ to $k$. At each step, we first identify the Advantage Particle Set (APS, shaded in red) which is a subset of the current PST $\Omega$ (blue ellipse), and then compute the current best solution $\pi_{k}^{\ast}$ (red dot) as the weighted average of the particles in APS. The PST is then moved to $\pi_{k}^{\ast}$ with the new axis length proportional to the vector $\bv = \pi_{k}^{\ast} - \pi_{k-1}^{\ast}$, thus evolving into (green ellipse). }}
\label{fig:vis_ro}
\end{figure}

%% file: mapping.tex
\subsection{Mapping of the Active Submap}
\label{sec:mapping}
Given the sequentially acquired RGB-D frames, the mapping process optimizes the network parameters of $f_s^\theta$ and $c_s^\lambda$ via minimizing the TSDF truncation-region and free-space losses, along with the RGB rendering loss. In this subsection, we focus on the mapping of the active submap and leave the refinement of inactive ones for the next subsection. Algorithm \ref{algo:pfopst} describes the mechanism of multi-submap  maintenance.

\jiazhao{\input{figure/algo_pfopst}}

\paragraph{Submap allocation}
The subvolume governed by a neural submap is an axis-aligned bounding box enveloping the viewing frustums (the far clipping plane is set to $5$m) of all keyframes. The subvolume of the active submap grows dynamically as new keyframes are being added. Whenever a new keyframe is selected, the subvolume is enlarged by a minimum expansion to enclose its viewing frustums. When any side of the subvolume reaches a predefined maximum length (set to $7$m), the subvolume stops expanding along that dimension. 
When the overlap between the viewing frustum of a keyframe and the subvolume is less than $75\%$ of the frustum \ffinal{(see \texttt{CheckOutBound()} in Algorithm \ref{algo:pfopst})}, a new submap is allocated and set as active and that keyframe is selected as its first/anchor keyframe. The previous active submap then becomes inactive.

\final{\paragraph{Submap initialization:} When a new submap is created, we perform initialization using its first keyframe shared with the previous active submap for $500$ epochs (found through experiments), to make sure that the MLP of the new submap is optimized sufficiently for a smooth transition of tracking across submaps} \ffinal{(see \texttt{CreateSubmap()} in Algorithm \ref{algo:pfopst})}.

\paragraph{Keyframe selection}
The selection of keyframes is via measuring the information gain of a frame \ffinal{(see \texttt{CheckKeyFrame()} in Algorithm \ref{algo:pfopst})}.
Based on the depth-to-TSDF loss, we compute for each frame an information gain used for filtering those frames which does not induce much novel information. Given a frame, we compute the depth-to-TSDF loss for each pixel: $\psi_s(\bT_s^{-1}(\bR_t\bx_p+\bt_t))$. If the proportion of pixels having a small error ($<0.05$) is lower than $65\%$, the frame is selected as a keyframe.
These thresholds were found through experiments and are then kept fixed.
To avoid selecting keyframes too frequently, we stipulate that the minimum spacing between two keyframes is $30$ frames \ffinal{(see \texttt{InserKeyFrame()} in Algorithm \ref{algo:pfopst})}.

\paragraph{Active submap optimization}
The optimization of the active submap \emph{at each frame} involves five different frames.
First of all, the current frame, after being tracked, always participates in map optimization.
The first/anchor keyframe of the active submap is also selected with its pose being fixed during optimization.
Fixing the anchor pose avoids free drifting of the entire submap.
Besides the above two frames, we randomly selected another three keyframes in between.
If there are fewer than three keyframes in between, already selected frames are duplicated up to five.
We found through experiments that using such five frames for optimization leads to a good balance between accuracy and efficiency. In summary, each frame participates in the optimization at least once, and more times if it is a keyframe.
The neural submap and the poses of the five frames (except for the anchor pose) are jointly optimized, which is essentially a local bundle adjustment for the active submap.
\ffinal{This local BA optimizes the following loss for $15$ iterations with a learning rate of $10^{-2}$ for submap update and $10^{-3}$ for pose optimization:
\begin{equation}\label{eq:fs-loss}
  \ffinal{\cL(\theta,\lambda,\bR_{i_1,\ldots,i_5},\bt_{i_1,\ldots,i_5}) = \sum_{b\in \cB}{\omega_\text{rgb} \cL_\text{rgb}^b + \omega_\text{fs} \cL_\text{fs}^b  + \omega_\text{tr} \cL_\text{tr}^b},}
\end{equation}
which sums up losses over the pixel batch set $\cB$ for all the involved five frames. The weights are: $\omega_\text{rgb}=1$, $\omega_\text{fs}=10$, and $\omega_\text{tr}=1000$.}


%% file: figure/algo_pfopst.tex
\IncMargin{0.5em}
\begin{algorithm}[t]
\caption{Mechanism of multi-submap maintenance}
\label{algo:pfopst}
\SetCommentSty{textsf}
\SetKwInOut{AlgoInput}{Input}
\SetKwInOut{AlgoOutput}{Output}
\SetKwFunction{BestState}{CompBestState}
\SetKwFunction{AxisLen}{CompAxisLength}
\SetKwFunction{SamplePST}{SamplePST}
\SetKwFunction{Move}{MovePST}
\SetKwFunction{Rescale}{RescalePST}
\SetKwFunction{InsertKF}{InsertKeyFrame}
\SetKwFunction{Outbound}{CheckOutBound}
\SetKwFunction{CreateSM}{CreateSubmap}
\SetKwFunction{Keyframe}{CheckKeyFrame}
\SetKwFunction{Init}{Initialize}
\AlgoInput{ RGB-D sequences $\{I^c_t, I^d_t\}$ and corresponding pose $\bx_t$ }
\AlgoOutput{ Submaps $M_j$ and corresponding keyframes $\{(I^c_i, I^d_i, \bx_i)\in\Omega(M_j)\}$ }
$M_0 \leftarrow$ \CreateSM{$I^c_0, I^d_0, \bx_0$} \tcp*{see \textit{submap allocation}}
$\Omega(M_0) \leftarrow$ \InsertKF{$I^c_0, I^d_0, \bx_0, \Omega(M_0)$}\;
$t \leftarrow 1$\; 
$j \leftarrow 0$\;
\Repeat{All frames are processed} { 
    \If {\Outbound{$\bx_t, M_{0:j}$}}{  
        $j \leftarrow j+1$\;
        $M_j \leftarrow$ \CreateSM{$\bx_t$}\; 
        $\Omega(M_j) \leftarrow$ \InsertKF{$I^c_t, I^d_t, \bx_t, \Omega(M_j)$}\;
    }
    
    \ForEach (\tcp*[f]{\small see \textit{keyframe selection} }){$ M_j \in \{M\} $}{
       \If {\Keyframe{$\bx_t, M_{j}$}}{ 
           $\Omega(M_j) \leftarrow$ \InsertKF{$I^c_t, I^d_t, \bx_t, \Omega(M_j)$}\;
        }
    }

    $t \leftarrow t+1$\;
}
\end{algorithm}
\DecMargin{0.5em} 

%% file: loop.tex
\subsection{Back-end Optimization and Loop Closure}
\label{sec:backend}
We create two threads running in parallel, one for the tracking and mapping of the active submap and one for the refinement of the inactive ones. This can improve the global map quality while ensuring realtime frame rate of online reconstruction.

\paragraph{Optimization of inactive submaps}
The inactive thread optimizes the inactive submaps sequentially and repeatedly. For the optimization of each inactive submap, we randomly select four keyframes belonging to the submap. The four keyframes, together with the first/anchor keyframe, are used to update the neural submap jointly. The poses of the four keyframes are also optimized with a small learning rate ($10^{-3}$). Such intra-submap local BA optimization is conducted for $10$ iterations \ffinal{(The number was determined through experiments for a trade-off between accuracy and efficiency).}

\final{\paragraph{Handling pose jump at submap revisiting}
When the camera moves into the subvolume of an inactive submap built previously, the inactive submap is re-activated. At this time, the overlapping keyframe, whose pose was just optimized in the last active submap, is now optimized again with the new active submap against its map built previously. Since the two maps may be misaligned due to drift, the pose of the keyframe may jump across the two optimizations. To avoid this jump, we first perform a local BA of the new active submap using this overlapping keyframe (together with other keyframes of it) for $10$ epochs, before starting tracking. This alignment of adjacent submaps makes the tracking transit smoothly.}

\paragraph{Loop detection and closure}
Since we concern about submap-level loop closure, loop detection is detected simply by checking whether the camera moves into the subvolume of an inactive submap. We detect only non-trivial loops involving at least four submaps. A more sophisticated scheme of loop detection can also be used.

To construct the optimization problem for loop closure, we first find for each pair of adjacent submaps (with subvolume overlap), $M_j$ and $M_k$, a set of point correspondences denoted as $\cC_{jk}$. To do so, we first identify the overlapping region between $M_j$ and $M_k$. In the overlapping region, we extract a set of surface points of $M_j$ at the zero level set of $\psi_j$, denoted by $\cS_j$. For each point $\bp^j_i \in \cS_j$, its correspondence on the surface of $\bq^k_i \in \cS_k$ can be found by first transforming it into the local coordinate frame of $M_k$ and then moving it along the gradient of $\psi_k$ for a distance of its TSDF value, similar to~\cite{fioraio2015large}:
\begin{equation}\label{eq:corr}
  \bq^k_i = \bT_k^{-1}\bT_j\bp^j_i - \psi_k(\bT_k^{-1}\bT_j\bp^j_i)\nabla\psi_k(\bT_k^{-1}\bT_j\bp^j_i),
\end{equation}
where $\nabla\psi$ denotes the gradient of TSDF field which is defined only within the truncation region. Therefore, our method can only find correspondences lying in the truncation regions. This suffices for our method since the drift between two consecutive submaps is usually quite small. Between the loop-closing submaps (the first and the last), however, the drift can be very large, for which we employ the existing feature detection and matching techniques~\cite{choi2012robust}. Given the correspondence $(\bp^j_i,\bq^k_i)\in \cC_{jk}$, we can formulate the following point-to-plane inter-submap pose constraint:
\begin{equation}\label{eq:error-p2p}
  e^{jk}_i = \left(\bp^j_i- \bT_j^{-1}\bT_k\bq^k_i\right)\cdot \bn^j_i,
\end{equation}
where $\bn^j_i = \nabla\psi_j(\bp^j_i)$ is the normal of $\bp^j_i$ in submap $M_j$.
In case two consecutive submaps have too low overlap such that their correspondence set is too small to pin down their relative transformation, we simply use a pose-to-pose constraint based on the tracked motion between the two submaps, i.e., $\bM_{s-1,s}$:
\begin{equation}\label{eq:error-pose}
  e^{s-1,s} = \log \left(\bM_{s-1,s}\bT_s^{-1}\bT_{s-1}\right),
\end{equation}
where $\log : SE(3) \rightarrow \mathfrak{se}(3)$ is the logarithmic map.
Putting the two constraints together, we solve for all submap poses by optimizing:
\begin{equation}\label{eq:ba-opt}
  \argmin_{\{\bT_1,\ldots,\bT_S\}} \sum_{j}\sum_{k}\sum_{i}\|e^{jk}_i\|^2 + \sum_{s}\|e^{s-1,s}\|
\end{equation}
The optimization is solved by Ceres~\cite{Ceres} with the Levenberg-Marquardt method.

\input{figure/submapreg}

%% file: figure/submapreg.tex
\begin{figure}[t]
\centering
\begin{overpic}
[width=0.9\linewidth]
{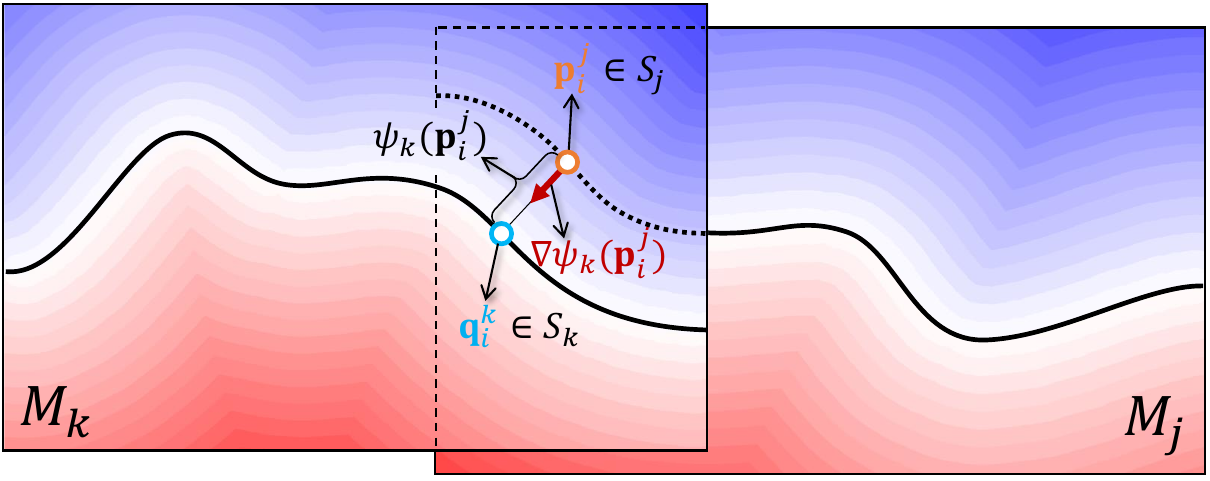}
\end{overpic}
\caption{
Finding correspondence for a pair of adjacent submaps. Given a surface point $\bp^j_i$ of one submap $M_j$, its correspondence is found by moving it along the gradient of \emph{the other} submap $M_k$, denoted by $\nabla\psi_k(\bp^j_i)$, for a distance of $\psi_k(\bp^j_i)$. Submap transformations are omitted for brevity.
}
\label{fig:submapreg}
\end{figure}

%% file: implement.tex

\section{Implementation details}
\label{sec:impl}

\paragraph{Parameter settings}
\ffinal{For efficiency, all losses are computed with downsampled $384$ pixels for both depth and RGB images following the sampling method of~\cite{zhang2021rosefusion} in which images are stripe downsampled into $16$ rows by $24$ columns, approximately $\frac{1}{30}$ of the original resolution which is $640\times480$.} We provide a detailed illustration to explain the downsampling algorithm in the supplemental material.
For each pixel, the sampling of 3D points on its ray is performed in two phases.
First, we uniformly sample $20$ points along the ray across the free space and the truncation region.
We then sample additional $20$ points uniformly within the truncation region.
In the second phase, we sample $10$ more points around the point having the smallest TSDF value.
Therefore, we sample $50$ points per ray in total.
For the tracking of active submap, RO is performed for $10$ iterations and GO for $10$ epochs.
The optimization of mapping also runs for $10$ epochs in each update.
For RO, $2048$ particles are pre-sampled and evolved; other parameter settings follow~\cite{zhang2020fusion}.
The batch size is $19,200$ $(384 \text{ pixels}\times50 \text{ points})$ for GO in tracking and $96,000$ $(384\text{ pixels}\times50\text{ points}\times5\text{ keyframes})$ for mapping.
\ffinal{All these numbers were found through experiments.}

\paragraph{GPU implementation}
The most time-consuming operation in a neural SLAM/reconstruction is the MLP training and inference based on all sample points in a batch. Furthermore, the fitness evaluation and filtering of particles in RO is also costly. To accelerate the computation, we make use of the Graph Execution mode of Tensorflow and compile all the core computations above into computational graphs. The computations are ``traced'' only once and can be called repeatedly and run efficiently in the GPU.

The optimization of the active and inactive submaps runs in separate processes in the GPU concurrently. Most of the time, the two processes work independently and they communicate with each other only when a new submap is created and the switching of active submaps happens. When the inactive submap adjacent to the active is being optimized, the two processes may update their overlapping keyframes jointly. To avoid ``dirty write'' of the overlapping keyframe residing in the shared memory, we set a write lock to ensure that it is optimized alternately by the active and inactive processes.



%% file: results.tex

\section{Results and evaluations}
\label{sec:result}
We provide both quantitative and qualitative results in this section.
\emph{Live demos can be found in the accompanying video.}

\subsection{Datasets and Metrics}
\label{sec:benchmark}

\paragraph{Datasets.}
We evaluated our method on three diverse public distastes, 
including
\DSREPL~\cite{straub2019replica},
\DSSN~\cite{dai2017scannet},
and \DSFCM~\cite{zhang2021rosefusion}.
\DSREPL is a synthetic dataset containing rendered (with noise added) RGB-D sequences.
\DSSN is a real dataset of captured RGB-D sequences.
\DSFCM is a challenging dataset of sequences with fast camera motions.
The dataset is composed of a synthetic part (\DSFCMS) and a real captured (\DSFCMR) part.
\DSFCMS is built with $10$ \DSREPL scenes. The RGB-D sequences have the linear speed of camera motion varying in $[1, 4]$ m/s and the angular speed in $[0.9, 2.2]$ rad/s, with synthesized motion blur effect for RGB images and depth noise for depth maps.
\DSFCMR contains $24$ real RGB-D sequences captured for $12$ scenes with fast camera motions (linear speed up to $5.47$m/s). For each sequence, a full and dense reconstruction scanned with a laser scanner is provided as ground truth for evaluating reconstruction accuracy and completeness.


\input{figure/tab_ablation}

To better evaluate the scalability of RGB-D reconstruction methods, we contribute a new real-world dataset of RGB-D sequences capturing six large-scale indoor scenes (with area up to $200$m$^2$), named FastCaMo-Large. The sequences were captured using an Azure Kinect DK under fast camera motions, and the individual size of each scene can be found in the supplemental material.

\paragraph{Evaluation metrics.}
When the ground-truth trajectory is available, we measure the camera tracking quality based on the Absolute Trajectory Error (ATE)~\cite{sturm2012benchmark}. To estimate ATE, the trajectory to be evaluated is first rigidly aligned to the ground truth. ATE is then estimated as the mean of pose differences of all frames.
We also measure the per-frame pose accuracy based on Translation Error (TE).
In addition, we use Relative Pose Error (RPE) to evaluate the relative pose differences over a fixed time interval between the estimated and the ground-truth trajectories. RPE is suited for evaluating local trajectory accuracy.
TE and RPE do not require a pre-alignment of the estimated and ground-truth trajectories. As long as the trajectories start from the same initial pose of the very first frame, they can always be estimated for the following frames in the reference system of the first frame.
For a fair comparison, we conducted multiple runs of our method and other open-sourced neural-based methods using different random seeds. Specifically, we executed each method five times and recorded the average result as the final outcome. This approach helps mitigate the impact of random variations and provides a more reliable and robust evaluation of performance.

To evaluate the reconstruction quality, we measure the reconstruction completeness and accuracy based on ground-truth surface reconstruction. Following~\cite{zhang2021rosefusion}, we measure completeness as the percentage of the inlier portion of the ground-truth surface and accuracy by RMS error of all reconstructed points against the ground-truth surface.

\input{figure/plot_rogo}

\subsection{Ablation Studies}
\label{sec:ablation}
We conduct a series of ablation studies to verify the necessity of the various key design choices of our method:
\begin{compactitem}
\item \textbf{No Classification (No C)}: TSDF prediction is implemented with regression as in existing methods.
\item \textbf{No RO}: The tracking optimization is performed by GO only as in existing methods.
\item \textbf{No GO}: There is no GO-based pose refinement after RO.
\item \textbf{No Uncertainty (No U)}: The fitness evaluation (\Eq{fitness}) in RO is not weighted by uncertainty.
\item \textbf{No Submap Initialization (No SI)}: No initialization is performed for newly allocated submaps.
\item \textbf{No Smooth Revisit (No SR)}: No handling of pose jump for smooth transition is done at submap revisiting.
\item \jiazhao{\textbf{No Loop Closure (No LC)}: No loop closure optimization of submaps is conducted.}
\end{compactitem}

The evaluation is conducted on 6 sequences from \DSSN and 4 from \DSFCMS. \Tab{ablation} compares the tracking accuracy (ATE) of our method and the various baselines.
It can be seen that ``No SI'' and ``No SR'' cause the most accuracy drop (and even failures) among all baselines, suggesting their importance to stable tracking. This also suggests that the handling of smooth transition between two submaps especially when revisiting an inactive submap is critical to the overall tracking quality. The combination of RO and GO produces better accuracy than either one of them. In particular, RO's effect is more prominent for fast-camera-motion sequences. For sequences with ordinary camera motions, GO's contribution seems more significant.
Classification is also influential in tracking accuracy as implied by the results of ``No C''.
Although relatively less significant, ``No U'' does affect the final tracking accuracy, hinting that the uncertainty estimated by the classification network output is indeed meaningful.
\rev{The effect of ``No LC'' manifests the necessity of our submap-level loop closure for fast-camera-motion sequences. Note, however, that the local BAs of inactive submaps are also turned off in ``No LC''. This is because the local BAs rely on globally consistent frame poses provided by the global optimization of loop closure.}

In \Fig{plot-rogo}, we show plots of tracking accuracy over iteration steps for RO only, GO only and our RO+GO. The tracking accuracy is measured by per-frame translation error averaged over all frames of \texttt{scene0207} of \DSSN. The full ranges of TE variation are depicted by bars and the ranges of medium half by boxes. Although RO finds a good solution efficiently and converges faster than GO, GO can be used to refine the solution found by RO, leading to a better solution. Indeed, RO excels at finding a good initial solution by getting rid of local minima due to its randomized nature. GO is good at finding a better optimum in the vicinity of an initial guess. Combining the two makes our method enjoy the advantages of both worlds.

\input{figure/plot_switch}

To demonstrate the effect of pose jump handling when revisiting an inactive submap, \Fig{plot-switch} plots of the tracking accuracy (RPE) over time with and without jump handling. When no jump handling is conducted, a spike of the RPE curve is observed, which also affect adversely the pose tracking of the following frames (see the higher RPEs of the following time stamps).

\subsection{Quantitative Comparisons}
We quantitatively evaluate our method against several state-of-the-art methods for both ordinary and fast-motion sequences.

\input{figure/tab_comp_rpl}

\paragraph{Comparison on \DSREPL}
\Tab{comp-rpl} compares ATE RMSE on 8 sequences of \DSREPL between our method and three state-of-the-art neural online RGB-D reconstruction methods (iMAP~\cite{sucar2021imap}, NICE-SLAM~\cite{zhu2022nice} and Vox-Fusion~\cite{yang2022vox}).
On these relatively easy sequences, our method achieves comparable accuracy to Vox-Fusion with much less running time and memory footprint (see \Sec{time-mem}). \rev{We also evaluate the reconstruction quality of our method in comparison to VoxFusion and NICE-SLAM. The results demonstrate that our method achieves comparable reconstruction quality to VoxFusion, while outperforming NICE-SLAM. \emph{The results can be found in the supplemental material.}}

\input{figure/tab_comp_sn}
\paragraph{Comparison on \DSSN}
\Tab{comp-sn} reports the comparison on 8 sequences of \DSSN (\jiazhao{index $"0"$ for each scene}).
These sequences include not only those tested by the alternatives~\cite{zhu2022nice,yang2022vox} in their papers but also new ones which we believe are more challenging. Our method achieves comparable accuracy to the best-performing method.
Sequences such as \texttt{scene0000}, \texttt{scene0181}, \texttt{scene0011} and \texttt{scene0059} contain complex camera trajectories (with multiple loops), on which our method demonstrates good results due to the back-end optimization.
The \texttt{scene0024} sequence is the most challenging one which contains large open areas and lacks geometric details. Our method works the best on this sequence due to the robust tracking method employed.

\input{figure/tab_comp_fcm}

\input{figure/tab_reconstuction_synth}
\input{figure/tab_comp_fcm_tm}

\paragraph{Comparison on \DSFCMS}
 \rev{\Tab{comp-fcm} reports a comparison on 10 sequences of \DSFCMS. All these sequences were recorded with fast camera motions. Among the methods, ours is the only one that can reconstruct all the sequences with decent accuracy. \texttt{Office\_3} is the most challenging one due to the large camera rotations involved, on which our method achieves an ATE of $17.4$cm. Our method is the first, to our knowledge, that realizes online neural RGB-D reconstruction under fast camera motions. Besides tracking accuracy, Table~\ref{tab:comp-recon-synth} compares the reconstruction quality. Our method exhibits the best completion and accuracy for most of the sequences.
We also compare our method with two traditional RGB-D reconstruction methods, i.e., BunldeFusion~\cite{Dai2017} and ElasticFusion~\cite{Whelan2015}. The results are reported in Table~\ref{tab:comp-fcm-tm}. Generally speaking, the performance of the current neural SLAM approaches is still not comparable to traditional ones. On the fast-camera-motion sequences, however, our method performs better than the two traditional methods, thanks to the integration of gradient-based and randomized optimizations in the neural setting and the efficient learning of submaps.}

\ffinal{
\paragraph{Comparison on \DSTUM}
Table ~\ref{tab:comp-tum} reports a comparison of tracking accuracy on three \DSTUM sequences with slow camera motions and small scene scales. Traditional methods (rows 5-8) are generally more accurate than neural-based ones (rows 1-4) with dedicated designs such as feature tracking and depth noise modeling.
Our method achieves comparable performance to NICE-SLAM.
The advantage of our method is more prominent for fast-motion and large-scale sequences.}

\input{figure/tab_comp_tum}


\subsection{Qualitative Results}
\label{sec:qualitative}

\input{figure/render_results.tex}

\rev{\paragraph{Visual comparison of neural rendering.} We provide the results of the neural rendering in \Fig{render-results} on sequences from \DSFCML, \DSSN, and \DSFCMS. Our method achieves higher rendering quality under challenging lighting conditions in real-world environments (rows 1-2). 
For the fast-motion sequences (row 4), NICE-SLAM finds difficulty in learning geometry and appearance within a short time interval, leading to suboptimal results. In contrast, our method consistently produces high-quality rendering outputs throughout the sequences (row 4).
On the quantitative side, our method outperforms NICE-SLAM by $51.3\%$ in PSNR on \DSFCML. \emph{Please refer to the supplemental material.}
}

\input{figure/room_scal_vis.tex}

\input{figure/loop_closure}

\paragraph{Visual comparison of reconstruction.} \rev{We compare the reconstruction quality of our method with several mainstream methods including both neural-based~\cite{yang2022vox, zhu2022nice} and traditional-based ones~\cite{Dai2017, Whelan2015}. The evaluation was performed on the \DSFCMR and \DSFCMS datasets and the results are shown in Figure~\ref{fig:fig-room-scale}. Note that our method achieves better reconstruction quality with fewer artifacts and more complete geometry. This is also reflected by the better trajectory conformance of our method against the ground truths.}


\rev{In \Fig{gallery}, we show a gallery of reconstruction results on several large-scale indoor scenes of the \DSFCML dataset. Here, we observe that our method exhibits significant advantages in terms of both completion and quality, especially in scenarios with large loops (columns 1 and 4). Our method attains higher quality thanks to 1) the RO-based pose optimization leading to robustness and 2) the classification-based design making the network lightweight and fast to learn. In most sequences, our method preserves geometric details better (see the zoom-in views). We attribute this to our distributed neural representation where each submap takes charge of only a local region and hence more scene details can be memorized.}

\paragraph{Visualization of loop closure.}
In \Fig{loop-closure}, we present a visualization of loop closure with two examples.
For the \texttt{Apartment\_II} sequence of \DSFCMR (top row), the trajectory starts to drift in the middle image. The trajectory is then corrected when a loop is formed and closed by the final submap shown in the right image.
The same goes for the sequence shown in the bottom row. After loop closure, the overall tracking error is minimized.
Since each submap has been well optimized with local BAs in both active and inactive processes, our method only needs to perform submap-level registrations to close a loop.
Thanks to the mechanism of the local-frame definition of a neural submap and pair-wise alignment of adjacent submaps, we can achieve submap-level loop closure straightforwardly and efficiently.
This flexibility is a clear advantage over the single-implicit-map~\cite{sucar2021imap} and the feature-grid-based approaches~\cite{zhu2022nice,yang2022vox} based on which it is hard to realize global map update caused by loop closure. \final{Examples for with and without non-trivial loops can be found in the supplemental material. We observe that our method consistently exhibits robustness across various types of loops.}

\input{figure/plot_memory}

\input{figure/tab_timing}

\subsection{Runtime and Memory Analysis}
\label{sec:time-mem}
\Tab{timing} reports the average runtime for one iteration of the various algorithmic components of our method tested on the \texttt{scene0000} sequence of \DSSN. The time was measured on a workstation with an Intel\textsuperscript{\textregistered} Core\textsuperscript{TM} i9-1290K CPU @ 3.9GHz $\times$ 16 with 32GB RAM and an Nvidia GeForce RTX 3090Ti GPU with 24GB memory.
In terms of total runtime for full reconstruction of the tested sequences, our method is $4\times$ faster than NICE-SLAM and $3\times$ faster than Vox-Fusion.

In \Fig{plot-memory}, we compare the average and maximum running memory cost of
NICE-SLAM, Vox-Fusion, and our method for increasing scene scales. Our
method leads to the smallest memory footprint and the cost for the three scenes. In fact, the main storage cost of our method is the sample batches for the optimization of the active submap and one of the inactive submaps being refined. Such memory cost does not increase drastically with growing scene scales. Our method does not require extra memory for storing feature grids as in the alternative methods.

\input{figure/gallery}


%% file: figure/tab_ablation.tex
\begin{table}[!t]\centering
\caption{
\jiazhao{Ablation study of seven design choices on tracking accuracy (ATE in cm) over 6 sequences of \DSSN (top rows) and 4 of \DSFCM (bottom rows). The best results for each sequence are highlighted in \best{blue} color. `--' indicates that the tracking failed for the corresponding method.}
}\vspace{-5pt}
\scalebox{0.95}{
\setlength{\tabcolsep}{0.7mm}
\begin{tabular}{l|c|c|c|c|c|c|c|c}
\hline
Method                  & No C    & No RO    & No GO   & No U   & No SI  & No SR  & No LC  & Full        \\ \hline\hline
\texttt{Scene0000}      & $11.5$  & $12.8$   & $19.9$  & $12.1$ & --     & $17.5$ & $27.5$  & \best{7.9} \\ \hline
\texttt{Scene0106}      & $10.8$  & $13.9$   & $15.3$  & $11.3$ & $17.3$ & $20.9$ & $35.5$  & \best{9.7} \\ \hline
\texttt{scene0169}      & $12.3$  & $15.1$   & $36.5$  & $13.5$ & --     & --     & --  & \best{9.7} \\ \hline
\texttt{scene0181}      & $14.9$  & $17.5$   & $29.6$  & $14.7$ & --     & $18.4$ & $15.1$  & \best{14.2} \\ \hline
\texttt{scene0207}      & $10.2$  & $8.9$    & $19.5$  & $8.2$  & $20.8$ & $19.5$ & --  & \best{7.8}  \\ \hline
\texttt{scene0011}      & $9.1$   & $14.2$   & $18.6$  & $7.9$  & --     & $10.1$ & $15.4$  & \best{7.5}  \\ \hline\hline
\texttt{Apartment\_1}   & $10.5$  & $27.6$   & $10.2$  & $11.0$ & --     & --     & $13.9$  & \best{7.0}  \\ \hline
\texttt{Hotel\_0}       & $7.3$   & $14.3$   & $6.2$   & $6.9$  & --     & $9.5$  & $10.3$  & \best{4.8}  \\ \hline
\texttt{Office\_0}      & $6.9$   & $19.1$   & $7.6$   & $6.8$  & --     & $6.8$  & $6.6$   & \best{3.6}  \\ \hline
\texttt{Room\_0}        & $8.1$   & $40.6$   & $8.9$   & $7.2$  & --     & $20.1$ & $28.0$  & \best{4.8}  \\ \hline
\end{tabular}
}
\label{tab:ablation}
\end{table} 

%% file: figure/plot_rogo.tex
\begin{figure}[t]
\centering
\begin{overpic}
[width=\linewidth]
{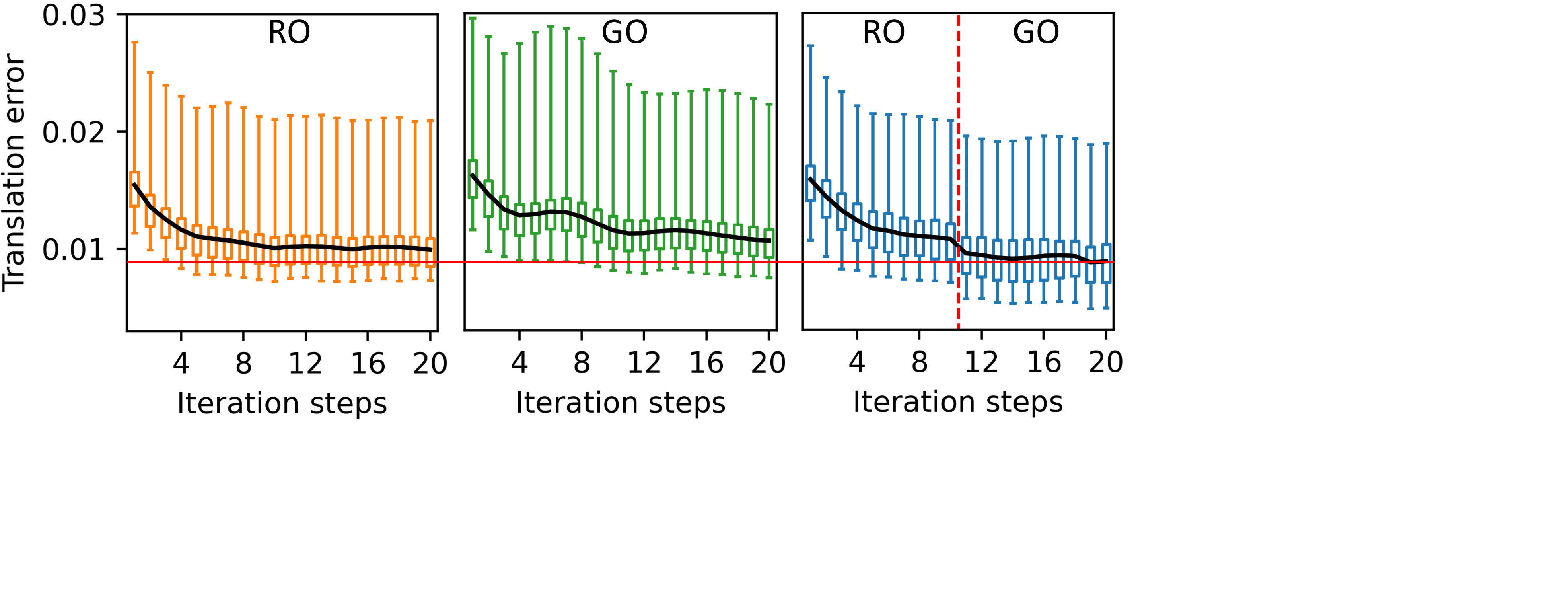}
    \put(18,37.5){\footnotesize (a) RO only.}
    \put(48,37.5){\footnotesize (b) GO only.}
    \put(78,37.5){\footnotesize (c) RO+GO.}
\end{overpic}
\caption{
Plots of per-frame tracking accuracy (TE) over iteration steps for (a) RO only, (b) GO only, and (c) RO and then GO (our method). Each plot shows TE averaged over all frames of a sequence (black curve) and ranges of variation (colored bars). In (c), the dashed red line indicates the switching point from RO to GO. The horizontal solid line across the three plots is drawn for a clear comparison of the converging error by the three methods.
}
\label{fig:plot-rogo}
\end{figure}

%% file: figure/plot_switch.tex
\begin{figure}[t]
\centering
\begin{overpic}
[width=\linewidth]
{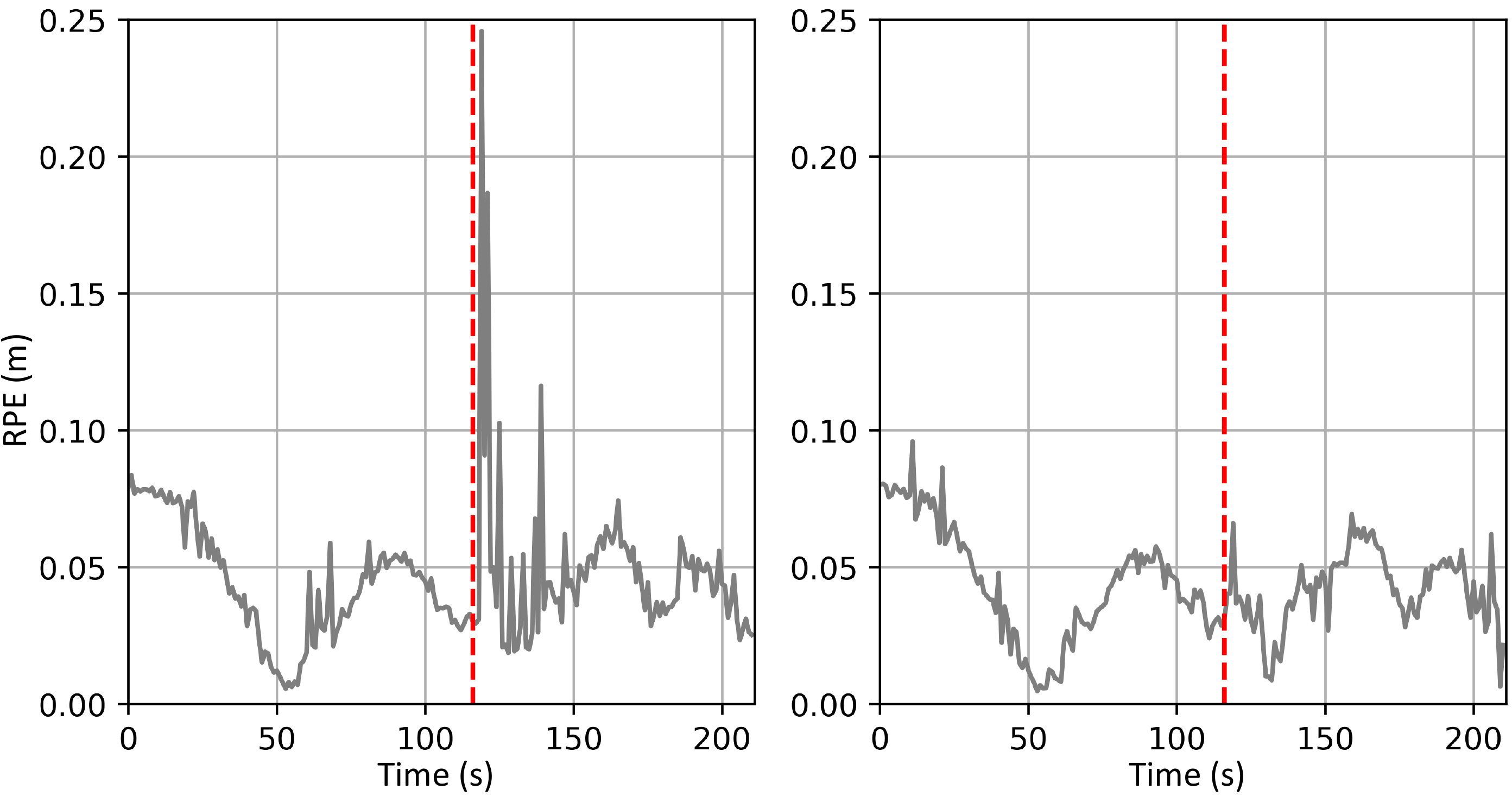}
    \put(16,53){\footnotesize (a) w/o jump handling}
    \put(66,53){\footnotesize (b) w/ jump handling}
\end{overpic}
\caption{
Plots of Relative Pose Error (RPE) over time with and without jump handling at submap revisiting. The red dashed line indicates the time at which an inactive submap is revisited and a spike is observed when no jump handling is conducted.
}
\label{fig:plot-switch}
\end{figure}

%% file: figure/tab_comp_rpl.tex
\begin{table}[!t]\centering
\caption{
\jiazhao{Comparing tracking accuracy (ATE RMSE in cm) on 8 RGB-D sequences of \textbf{\DSREPL}. The best and the second best results for each sequence are highlighted in \best{blue} and \sbest{green} colors, respectively.}
}\vspace{-5pt}
\scalebox{0.95}{
\setlength{\tabcolsep}{1.5mm}{
\begin{tabular}{l|c|c|c|c}
\hline
Sequence                & iMAP    & NICE-SLAM & Vox-Fusion & MIPS-Fusion  \\ \hline\hline
\texttt{Room-0}         & $70.1$  & $1.7$     & \best{0.3}      & \sbest{1.1}           \\ \hline
\texttt{Room-1}         & $4.5$   & $2.0$     & \sbest{1.3}      & \best{1.2}           \\ \hline
\texttt{Room-2}         & $2.2$   & $1.6$     & \best{0.5}      & \sbest{1.1}           \\ \hline
\texttt{Office-0}       & $2.3$   & \sbest{1.0}     & \best{0.7} & \best{0.7}           \\ \hline
\texttt{Office-1}       & $1.7$   & \sbest{0.9}     & $1.1$      & \best{0.8}          \\ \hline
\texttt{Office-2}       & $4.9$   & $1.4$     & \best{0.5} & \sbest{1.3}         \\ \hline
\texttt{Office-3}       & $58.4$  & $4.0$     & \best{0.3} & \sbest{2.2}           \\ \hline
\texttt{Office-4}       & $2.6$   & $3.1$     & \best{0.6} & \sbest{1.1}           \\ \hline
\end{tabular}
}}
\label{tab:comp-rpl}
\end{table} 

%% file: figure/tab_comp_sn.tex

\begin{table}[!t]
\centering
\caption{Comparing tracking accuracy (ATE RMSE in cm) on 8 RGB-D sequences of \textbf{\DSSN}. The best and the second best results for each sequence are highlighted in \best{blue} and \sbest{green} colors, respectively. '--' indicates that the tracking failed for the corresponding method.}
\label{tab:comp-sn}
\vspace{-5pt}
\scalebox{0.95}{
\setlength{\tabcolsep}{1.5mm}
\begin{tabular}{l|c|c|c|c}
\hline
Sequence           & iMAP  & NICE-SLAM   & Vox-Fusion  & MIPS-Fusion  \\ \hline\hline
\texttt{scene0000} & $--$  & \sbest{12.1} & 17.6        & \best{7.9}   \\ \hline
\texttt{scene0106} & 17.5  & \sbest{9.2}& \best{8.8}   & 9.7          \\ \hline
\texttt{scene0169} & $--$  & \sbest{11.2} & 20.0        & \best{9.7}   \\ \hline
\texttt{scene0181} & 32.1  & \best{13.9} & 19.0        & \sbest{14.2} \\ \hline
\texttt{scene0207} & 11.9  & \best{6.2}  & \sbest{7.5}  & 7.8          \\ \hline
\texttt{scene0011} & 18.4  & 8.2         & \best{7.4}   & \sbest{7.5}  \\ \hline
\texttt{scene0024} & $--$  & $--$        & \sbest{31.0} & \best{7.8}   \\ \hline
\texttt{scene0059} & $--$  & \sbest{12.8} & 35.5        & \best{10.7}   \\ \hline
\end{tabular}
}
\end{table}

%% file: figure/tab_comp_fcm.tex
\begin{table}[!t]\centering
\caption{
\jiazhao{Comparing tracking accuracy (ATE RMSE in cm) on 10 fast-camera-motion RGB-D sequences of \textbf{\DSFCMS} (noise-free). The best and the second best results for each sequence are highlighted in \best{blue} and \sbest{green} colors, respectively. `--' indicates that the tracking was failed for the corresponding method.}
}\vspace{-5pt}
\scalebox{0.95}{
\setlength{\tabcolsep}{1.0mm}{
\begin{tabular}{l|c|c|c|c}
\hline
Sequence                    & iMAP & NICE-SLAM & Vox-Fusion & MIPS-Fusion  \\ \hline\hline
\texttt{Apartment\_1}       & --   & --        & \sbest{9.1}        & \best{7.0}           \\ \hline
\texttt{Apartment\_2}       & --   & --        & \sbest{4.1}         & \best{1.5}           \\ \hline
\texttt{Frl\_apartment\_2}   & --   & --        & \sbest{7.2}         & \best{1.9}           \\ \hline
\texttt{Hotel\_0}           & $20.3$   & \best{4.2}   & $5.0$         & \sbest{4.8}           \\ \hline
\texttt{Office\_0}          & $39.2$   & $8.4$   & \sbest{4.8}        & \best{3.6}          \\ \hline
\texttt{Office\_1}          & --  & $13.7$  & \best{4.6}        & \sbest{5.6}           \\ \hline
\texttt{Office\_2}          & --   & --        & \sbest{10.2}        & \best{7.4}           \\ \hline
\texttt{Office\_3}          & --   & \best{14.3}    & --        & \sbest{17.4}           \\ \hline
\texttt{Room\_0}            & --   & --      & \sbest{8.2}      &  \best{4.4}     \\ \hline
\texttt{Room\_0}            & --   & $29.7$    & \sbest{5.8}         & \best{5.1}          \\ \hline
\end{tabular}
}}
\label{tab:comp-fcm}
\end{table} 

%% file: figure/tab_reconstuction_synth.tex
\begin{table}[!t]\centering
\caption{
\jiazhao{Comparing reconstruction quality (completeness and accuracy) on 10 fast-camera-motion RGB-D sequences of \textbf{\DSFCMS} (noise-free). The best results for each sequence are highlighted in \best{blue}. `--' indicates that the tracking failed for the corresponding method.}
}
\vspace{-5pt}
\scalebox{0.87}{
\setlength{\tabcolsep}{0.7mm}{
\begin{tabular}{l|cc|cc|cc}
\hline
                & \multicolumn{2}{c|}{NICE-SLAM}  & \multicolumn{2}{c|}{Vox-Fusion} & \multicolumn{2}{c}{Ours}        \\ \cline{2-7} 
 &
  \multicolumn{1}{c|}{Compl.($\uparrow$)} &
  Acc.($\downarrow$) &
  \multicolumn{1}{c|}{Compl.($\uparrow$)} &
  Acc.($\downarrow$) &
  \multicolumn{1}{c|}{Compl.($\uparrow$)} &
  Acc.($\downarrow$) \\ \hline \hline
Apartment\_1     & \multicolumn{1}{c|}{-}    & -   & \multicolumn{1}{c|}{63.4} & 4.8 & \multicolumn{1}{c|}{\best{73.9}} & \best{4.2} \\ \hline
Apartment\_2     & \multicolumn{1}{c|}{-}    & -   & \multicolumn{1}{c|}{\best{93.1}} & \best{2.4} & \multicolumn{1}{c|}{64.8} & 4.9 \\ \hline
Frl\_aparment\_2 & \multicolumn{1}{c|}{-}    & -   & \multicolumn{1}{c|}{62.3} & 5.1 & \multicolumn{1}{c|}{\best{78.0}} & \best{4.3} \\ \hline
Hotel\_0        & \multicolumn{1}{c|}{84.4} & 3.9 & \multicolumn{1}{c|}{66.0} & 4.6 & \multicolumn{1}{c|}{\best{88.8}} & \best{3.4} \\ \hline
Office\_0       & \multicolumn{1}{c|}{92.9} & \best{2.9} & \multicolumn{1}{c|}{44.8} & 6.5 & \multicolumn{1}{c|}{\best{94.2}} & \best{2.9} \\ \hline
Office\_1       & \multicolumn{1}{c|}{53.8} & 5.7 & \multicolumn{1}{c|}{\best{72.3}} & 5.9 & \multicolumn{1}{c|}{67.9} & \best{4.3} \\ \hline
Office\_2       & \multicolumn{1}{c|}{-}    & -   & \multicolumn{1}{c|}{51.6} & 6.3 & \multicolumn{1}{c|}{\best{62.7}} & \best{4.8} \\ \hline
Office\_3       & \multicolumn{1}{c|}{\best{63.4}} & \best{4.9} & \multicolumn{1}{c|}{-}    & -   & \multicolumn{1}{c|}{44.4} & 6.4 \\ \hline
Room\_0         & \multicolumn{1}{c|}{-}    & -   & \multicolumn{1}{c|}{37.6} & 7.0 & \multicolumn{1}{c|}{\best{65.6}} & \best{4.8} \\ \hline
Room\_1         & \multicolumn{1}{c|}{65.3} & 4.8 & \multicolumn{1}{c|}{37.7} & 7.0 & \multicolumn{1}{c|}{\best{83.4}} & \best{3.4} \\ \hline
\end{tabular}
}
}
\label{tab:comp-recon-synth}
\end{table} 

%% file: figure/tab_comp_fcm_tm.tex
\begin{table}[!t]\centering
\caption{
\jiazhao{Comparing tracking accuracy (ATE RMSE in cm) on 10 fast-camera-motion RGB-D sequences of \textbf{\DSFCMS} (with noise). The best  results for each sequence are highlighted in \best{blue} color. `--' indicates that the tracking was failed for the corresponding method.}
}\vspace{-5pt}
\scalebox{0.95}{
\setlength{\tabcolsep}{1.0mm}{
\begin{tabular}{l|c|c|c}
\hline
Sequence                     & ElasticFusion & BundleFusion & MIPS-Fusion  \\ \hline\hline
\texttt{Apartment\_1}         & 40.9   & \best{4.6}        & 6.6           \\ \hline
\texttt{Apartment\_2}         & 40.7   & \best{2.2}         & 3.1           \\ \hline
\texttt{Frl\_apartment\_2}    & 43.8   & 83.6         & \best{2.6}           \\ \hline
\texttt{Hotel\_0}             & 22.3   & \best{2.7}         & 5.2           \\ \hline
\texttt{Office\_0}            & \best{2.3}    & 17.3        & 7.6          \\ \hline
\texttt{Office\_1}            & -   & -        & \best{17.4}           \\ \hline
\texttt{Office\_2}            & -   & -        & \best{24.9}           \\ \hline
\texttt{Office\_3}            & 43.8   & -        & \best{6.0}           \\ \hline
\texttt{Room\_0}              & -      & 8.2      &  \best{4.4}     \\ \hline
\texttt{Room\_0}              & 31.0   & 5.8         & \best{3.6}          \\ \hline
\end{tabular}
}}
\label{tab:comp-fcm-tm}
\end{table} 

%% file: figure/tab_comp_tum.tex
\begin{table}[!t]\centering
\caption{
Comparing tracking accuracy (ATE in cm) on three RGB-D sequences of \textbf{\DSTUM}. The best results for each sequence are highlighted in \best{blue}. 
}\vspace{-5pt}
\scalebox{0.95}{
\setlength{\tabcolsep}{2.5mm}{
\begin{tabular}{l|c|c|c}
\hline
Method      & fr1/desk       & fr2/xyz        & fr3/office     \\ \hline \hline
iMap        & 4.9cm          & 2.0cm          & 5.8cm          \\ \hline
DI-Fusion   & 4.4cm          & 2.4cm          & 15.6cm         \\ \hline
NICE-SLAM   & \best{2.7cm} & 1.8cm          & \best{3.0cm} \\ \hline
MIPSFusion  & 3.0cm          & \best{1.4cm} & 4.6cm          \\ \hline \hline
BAD-SLAM    & 2.3cm          & 2.2cm          & 2.3cm          \\ \hline
Kintinuous  & 2.0cm          & 1.1cm          & 1.7cm          \\ \hline
ORB-SLAM2   & 1.6cm          & \best{0.4cm} & 1.0cm          \\ \hline
\cite{Cao2018RealtimeHT}            & 1.5cm          & 0.6cm          & 0.9cm          \\ \hline
HRBF-Fusion & \best{1.4cm} & 0.5cm          & \best{0.7cm} \\ \hline
\end{tabular}
}}
\label{tab:comp-tum}
\end{table}

%% file: figure/render_results.tex
\begin{figure}[t]
\centering
\begin{overpic}
[width=0.95\linewidth]
{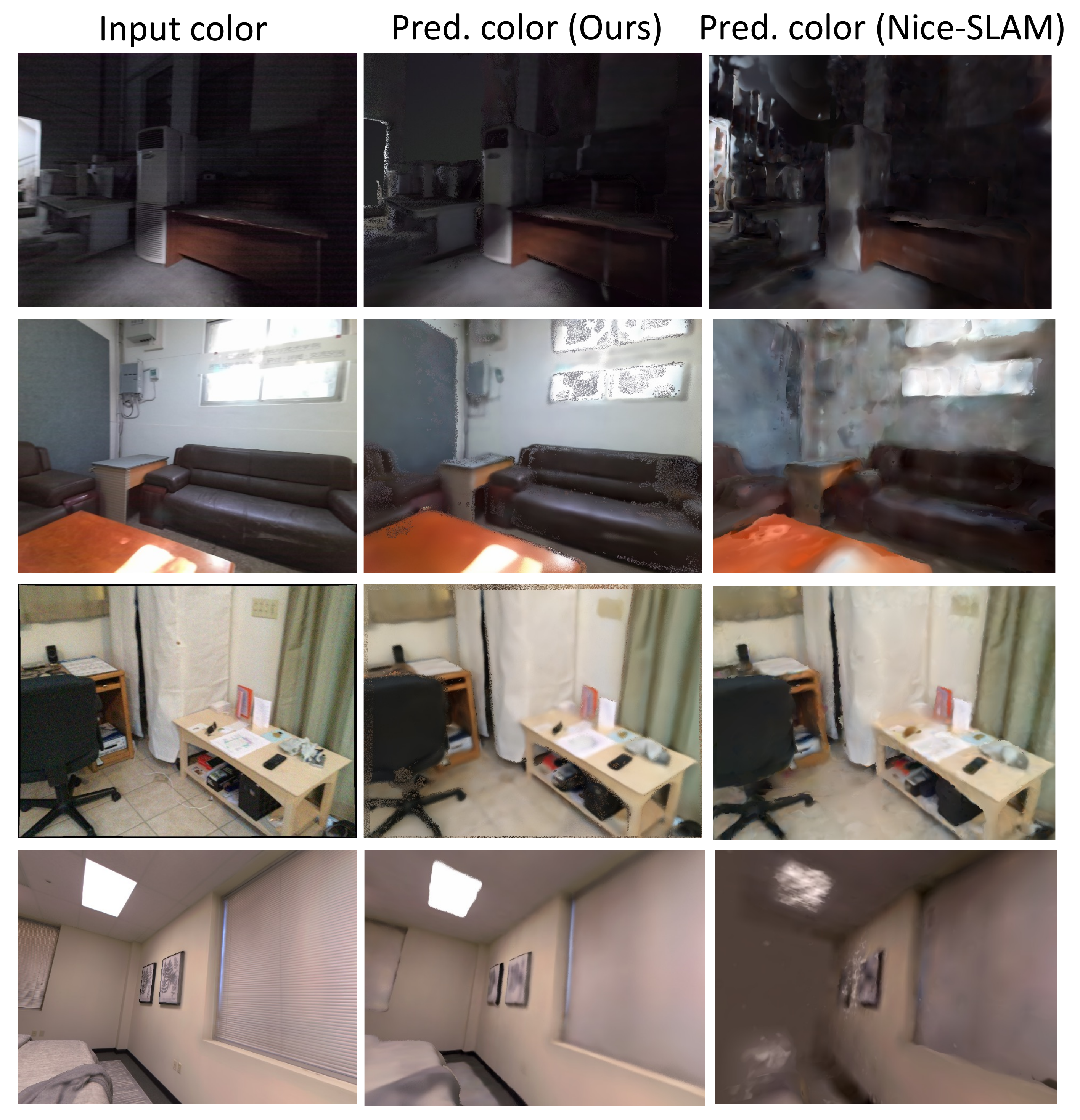}
\end{overpic}
\caption{
\rev{Rendering results of ours and NICE-SLAM on the sequences of \DSFCML (rows 1-2), ScanNet (row 3), and \DSFCMS (row 4).
}
}
\label{fig:render-results}
\end{figure}

%% file: figure/room_scal_vis.tex
\begin{figure}[t]
\centering
\begin{overpic}
[width=\linewidth]
{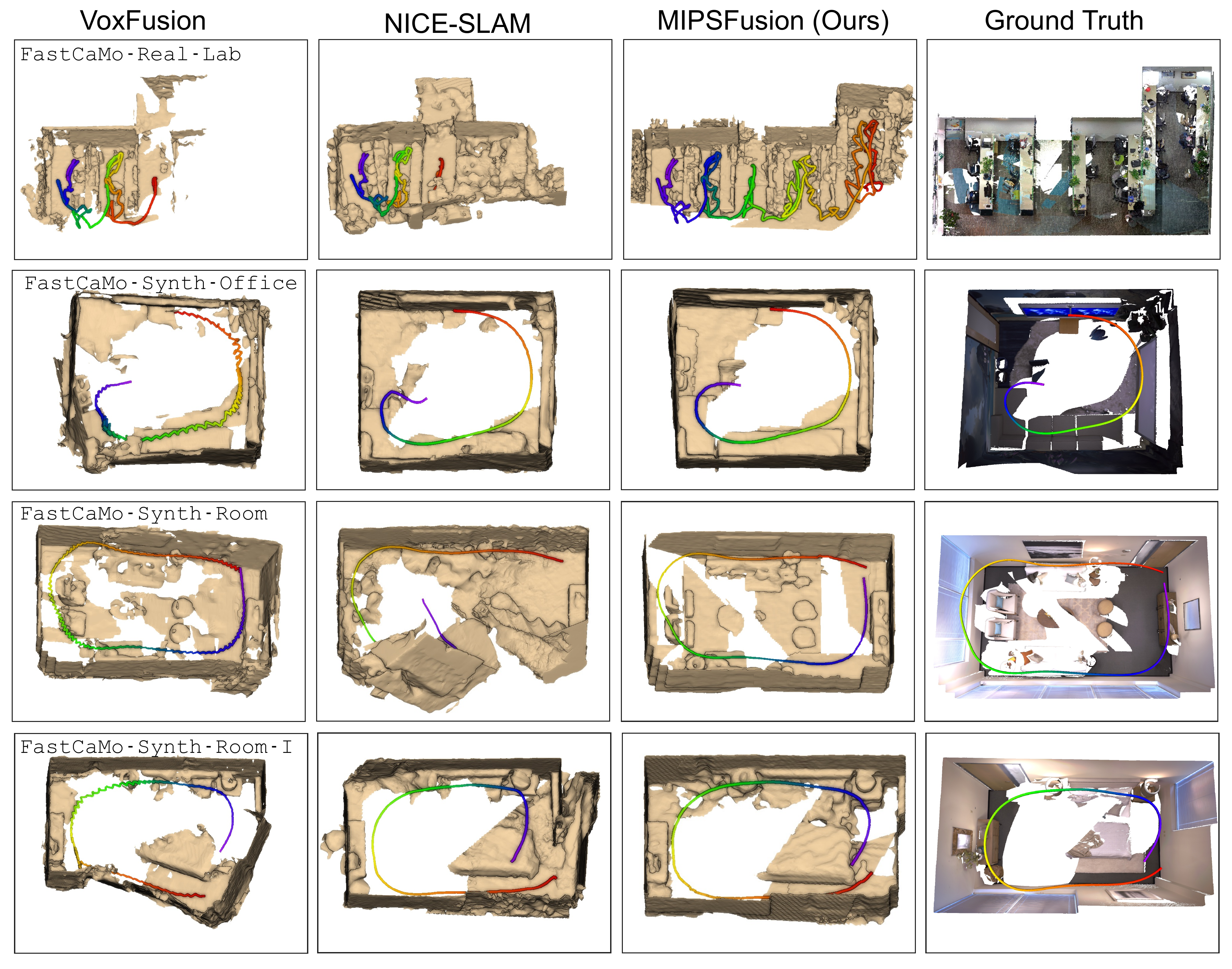}
\end{overpic}
\caption{
\jiazhao{Comparision of 3D reconstruction on four room-scale indoor scene sequences of FastCaMo-Real and FastCaMo-Synth.}
}
\label{fig:fig-room-scale}
\end{figure}

%% file: figure/loop_closure.tex
\begin{figure}[t]
\begin{overpic}
[width=0.9\linewidth]
{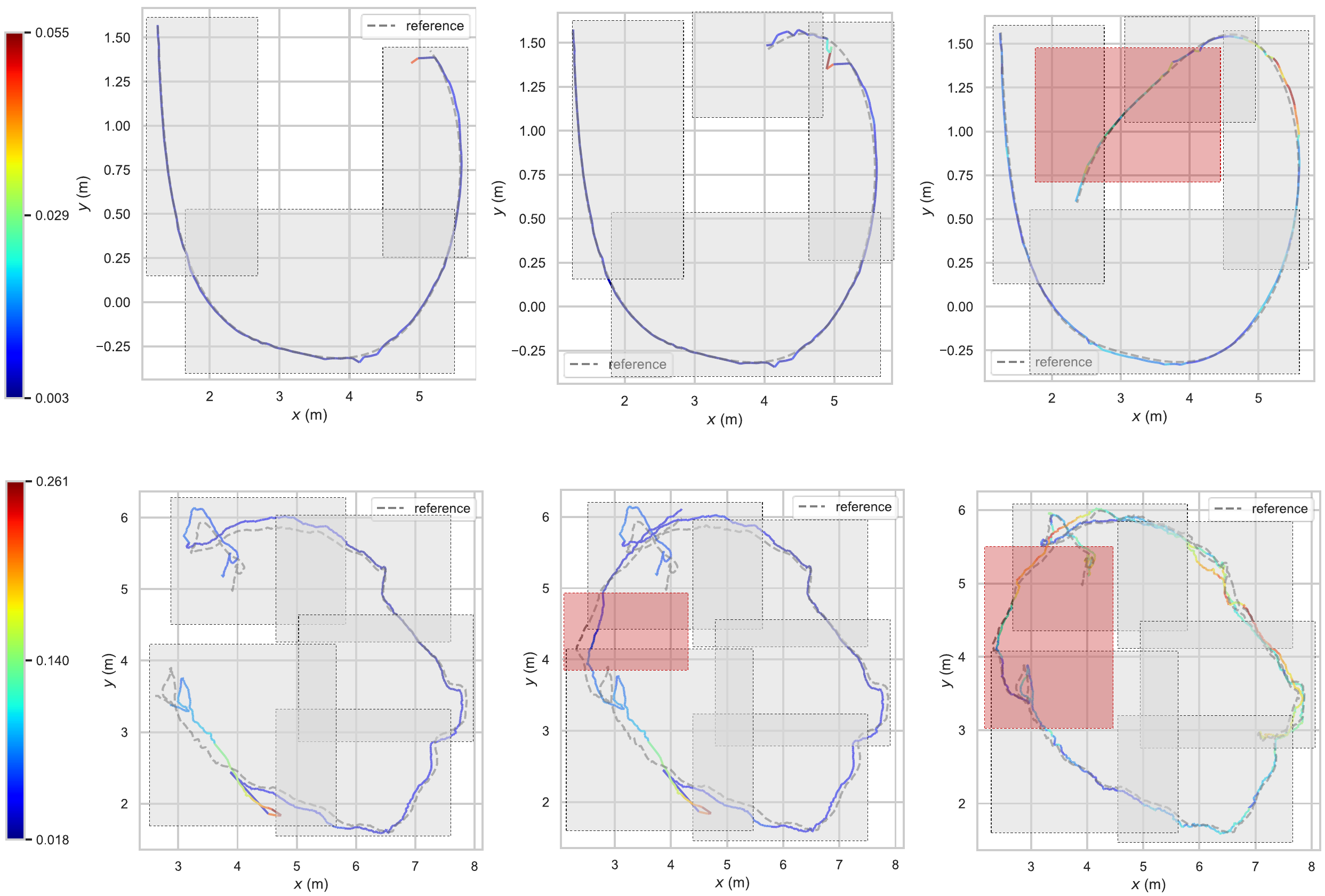}
    \put(32,68){\small \DSFCMR -- \texttt{Apartment\_II}}
    \put(35,32){\small \DSSN -- \texttt{Scene0169}}
\end{overpic}
\caption{
Visualization of loop closure on two sequences. Camera trajectories are visualized with solid curves and ground-truth reference with dashed curves. The tracking error is color-coded. Submaps are shown as grey boxes. The closing-loop submaps are shaded in red.
}
\label{fig:loop-closure}\vspace{-10pt}
\end{figure}

%% file: figure/plot_memory.tex
\begin{figure}[t]
\centering
\begin{overpic}
[width=\linewidth]
{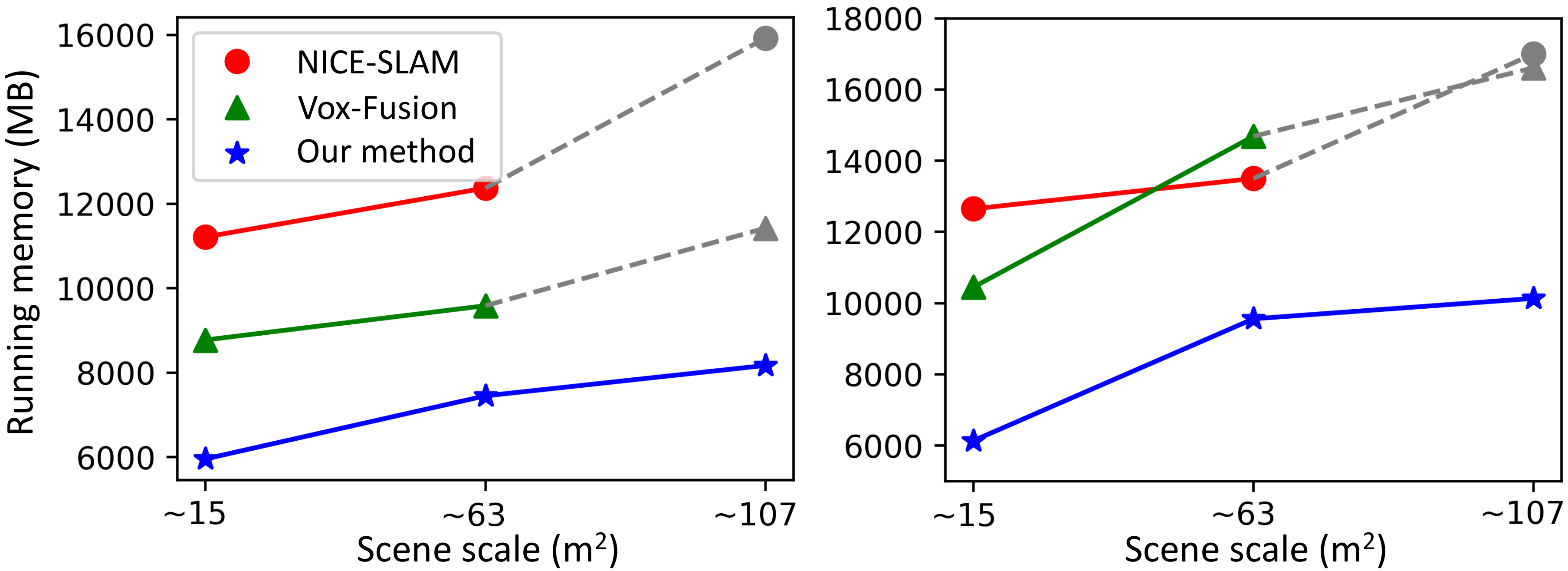}
    \put(14,36.5){\footnotesize (a) Average running memory.}
    \put(61,36.5){\footnotesize (b) Maximum running memory.}
\end{overpic}
\caption{
Comparing the average and maximum running memory cost of NICE-SLAM, Vox-Fusion and our method for increasing scene scales. Our method consumes the least running memory and the cost increases slowly with scene scale. The grey dots mean that the two alternative methods failed on those sequences.
}
\vspace{-4mm}
\label{fig:plot-memory}
\end{figure}

%% file: figure/tab_timing.tex
\begin{table}[!t]\centering
\caption{
Running time of various algorithmic components of our method profiled on a per-iteration basis.
}\vspace{-5pt}
\scalebox{0.95}{
\setlength{\tabcolsep}{1.5mm}{
\begin{tabular}{l|c}
\hline
Sequence                & Time     \\ \hline\hline
Tracking (RO)           & 4 ms       \\ \hline
Tracking (GO)           & 8 ms       \\ \hline
Mapping (1 frame)       & 9 ms       \\ \hline
Mapping (5 frames)      & 48 ms       \\ \hline
Create new submap       & 4 ms       \\ \hline
\end{tabular}
}}
\vspace{-4mm}
\label{tab:timing}
\end{table}

%% file: figure/gallery.tex
\begin{figure*}[t]
\centering
\begin{overpic}
[width=\linewidth]
{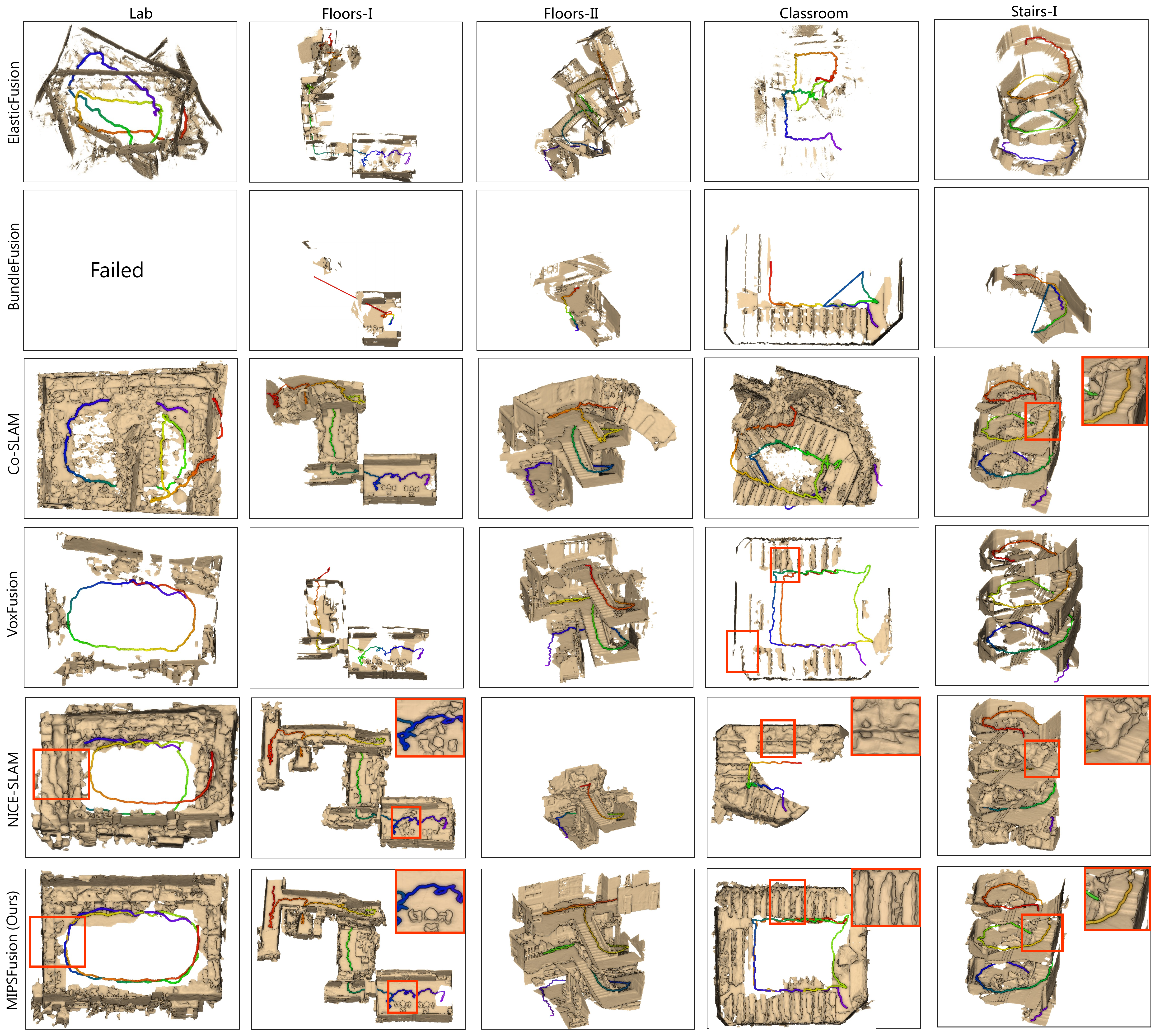}
\end{overpic}
\caption{
\yijie{Comparison of 3D reconstruction results by NICE-SLAM, Vox-Fusion, Co-SLAM, BundleFusion, ElasticFusion, and our MIPS-Fusion over five scenes from the \DSFCML datasets. Our method achieves better accuracy and completeness compared to the alternatives.}
}

\label{fig:gallery}\vspace{-10pt}
\end{figure*}

%% file: conclusion.tex

\section{Discussion and conclusions}
\label{sec:future}
With our work, we wish to bring it to the community's attention the potential of grid-free, purely neural representation for scalable and robust online RGB-D reconstruction. Our main design philosophy is two-fold. First, we adopt a flexible divide-and-conquer mapping scheme. Each submap, representing a subscene compactly, can be learned efficiently and refined distributively. The high-quality submaps together constitute a decent full reconstruction of the whole scene with submap-level global pose optimization. We believe that this mapping scheme has accomplished a good trade-off between flexibility and scalability. Second, we propose a hybrid tracking scheme in which randomized optimization is made possible based on two new designs on tracking loss. This enables efficient and robust tracking even under fast camera motions.

\paragraph{Limitations}
Our method has several limitations.
\emph{Firstly}, our tracking and mapping depends heavily on depth. When the depth input is of low quality, the reconstruction quality is unsatisfactory.
\emph{Secondly}, our loop detection is still simplistic. A loop may happen when the camera looks at a previously visited point without actually entering any inactive submap, which will be missed by our detection.
\emph{Thirdly}, when aligning two submaps having significant misalignment, robust feature detection and matching is still needed.
\emph{Finally}, our method does not handle view-dependent appearance such as specular since it does not model view directions in the neural radiance field as most existing works~\cite{sucar2021imap,zhu2022nice}.

\paragraph{Future works}
We expect that our work will inspire a rich set of future directions:
\begin{itemize}
  \item \rev{How to achieve a smarter submap allocation to ensure a better match between learning capacity and scene complexity? This may need a method for probing the representation forgetting~\cite{davari2022probing} of a neural submap against the acquired data.}
  \item \rev{How to realize end-to-end trainable loop detection and closure in one framework based on our MIPS representation? For example, it might be interesting to investigate an efficient neural remapping of submaps during loop closure based on a fast $SE(3)$-transformation of neural implicit maps~\cite{yuan2022algorithm}.}
  \item How to integrate the geometric and the photometric losses in a more principled way? Specifically, how to bridge and switch smoothly between the two is worth of investigating.
  \item How to fuse multi-modal input, e.g., inertial measurement, into online neural reconstruction using a similar technique to~\cite{zhang2022asro}?
  \item It seems a natural application to use neural submap representation for distributive and collaborative reconstruction of large scenes with a collection of robots~\cite{dong2019multi}.
  \item Another promising and interesting direction is to enhance neural submap representation for semantic scene reconstruction~\cite{zhang2020fusion,vora2021nesf}.
\end{itemize} 